\let\mypdfximage\pdfximage
\def\pdfximage{\immediate\mypdfximage}

\def\year{2021}\relax
\documentclass[letterpaper]{article} %
\usepackage{aaai21}  %
\usepackage{times}  %
\usepackage{helvet} %
\usepackage{courier}  %
\usepackage[hyphens]{url}  %
\usepackage{graphicx} %
\usepackage{natbib}  %
\usepackage{caption} %
\nocopyright %
\title{Evaluating Robustness to Context-Sensitive Feature Perturbations \\ of Different Granularities}

\author{
	Isaac Dunn, Laura Hanu\textsuperscript{\rm *}, Hadrien Pouget\textsuperscript{\rm *}, Daniel Kroening, Tom Melham
}

\affiliations {
	\\
	University of Oxford \\
	\textsuperscript{\rm *}Equal contribution \\
	\url{isaac.dunn@cs.ox.ac.uk}\\
}
\usepackage{microtype}
\usepackage{color}
\usepackage[utf8]{inputenc} %
\usepackage[T1]{fontenc}
\usepackage{amsmath}
\usepackage{amsfonts}
\usepackage{nicefrac}       %
\graphicspath{{./images/inkscape/}{./images/}}

\usepackage{collcell} %
\usepackage{xstring} %
\usepackage{booktabs} %
\usepackage{multirow}
\usepackage{array}
\usepackage{tabularx}

\usepackage{subcaption}

\newcommand{\backtick}{`}

\usepackage{amsmath,amsfonts,bm}

\def\eqref#1{equation~\ref{#1}}

\def\1{\bm{1}}

\def\eps{{\epsilon}}

\def\vp{{\bm{p}}}

\DeclareMathAlphabet{\mathsfit}{\encodingdefault}{\sfdefault}{m}{sl}
\SetMathAlphabet{\mathsfit}{bold}{\encodingdefault}{\sfdefault}{bx}{n}

\def\sA{{\mathbb{A}}}

\def\sX{{\mathbb{X}}}
\def\sY{{\mathbb{Y}}}
\def\sZ{{\mathbb{Z}}}

\newcommand{\R}{\mathbb{R}}

\begin{document}

\maketitle

\begin{abstract}

We cannot guarantee that training datasets
are representative of the distribution of
inputs that will be encountered during deployment.
So we must have confidence that our models do not
over-rely on this assumption.
To this end, we introduce
a new method that
identifies context-sensitive feature perturbations
(e.g.~shape, location,
texture, colour)
to the inputs
of image classifiers.
We produce
these changes by performing small
adjustments to the activation values
of different layers of
a trained generative neural network.
Perturbing at layers earlier in the generator
causes changes to coarser-grained features; perturbations
further on cause finer-grained changes.
Unsurprisingly, we find that state-of-the-art classifiers
are not robust to any such changes.
More surprisingly,
when it comes to coarse-grained feature changes,
we find that adversarial training against
pixel-space perturbations is not just unhelpful: it
is \emph{counterproductive}.

\end{abstract}

\section{Introduction}

\begin{figure*}[!h]
	\centering
	\begin{subfigure}{.26\textwidth}
		\centering
		\includegraphics[width=\linewidth]{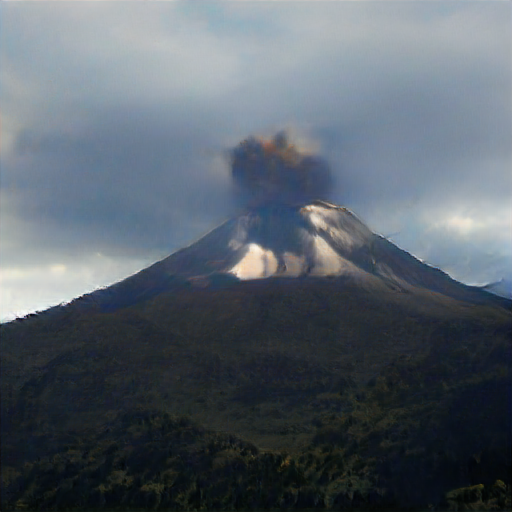}
		\caption{Original image.}
		\label{fig:sub1}
	\end{subfigure}\hfill
	\begin{subfigure}{.26\textwidth}
		\centering
		\includegraphics[width=\linewidth]{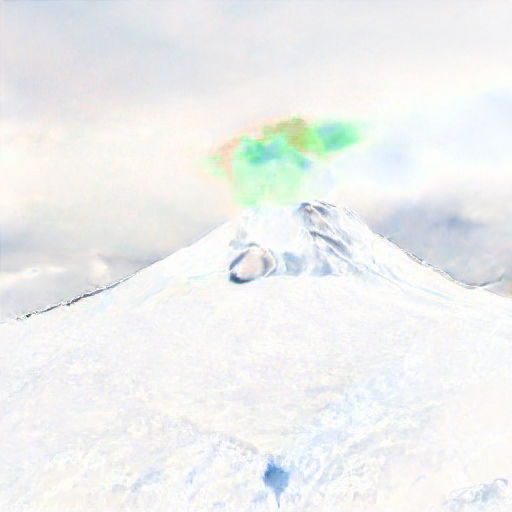}
		\caption{Difference from perturbation.}
		\label{fig:sub2}
	\end{subfigure}\hfill
	\begin{subfigure}{.26\textwidth}
		\centering
		\includegraphics[width=\linewidth]{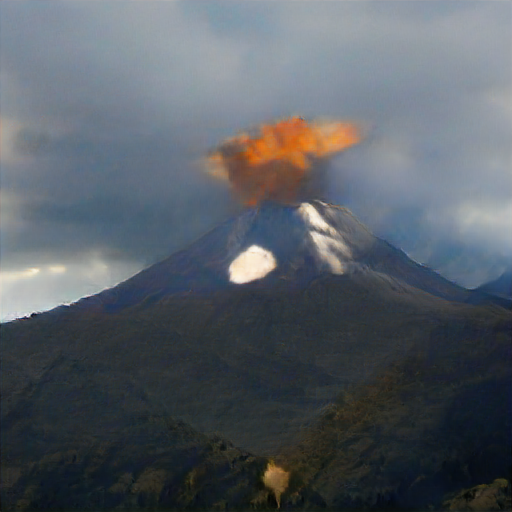}
		\caption{Perturbed image.}
		\label{fig:sub3}
	\end{subfigure}
	\caption{An example of changing
		the computed classification
		from `volcano' to target label `goldfish'
		using context-sensitive feature perturbations
		of all granularities.
		Coarser-grained changes include
		darkening the sky,
		causing an eruption of lava, and
		adding a rocky
		outcrop in the foreground;
		finer-grained changes include
		slightly flattening the curve of the volcano,
		and adjustments to the texture of the trees, rocks and cloud.
	}
	\label{fig:example_perturbation}
\end{figure*}

Deep learning models have proven to be 
powerful tools for tasks including image classification 
\cite{DBLP:journals/corr/abs-2003-08237,
DBLP:journals/corr/abs-1911-04252},
with the ability to automatically
identify useful features of
their training images and combine
these to provide accurate label predictions \cite{olah2020zoom}.
Under the assumption that data are
independently and identically distributed (i.i.d.),
they have shown a remarkable ability to generalise
to unseen inputs~\cite{DBLP:conf/nips/NeyshaburBMS17}.
However, it is increasingly clear
that the performance of these models
drops drastically without this
assumption;
optimising for
i.i.d.~accuracy alone results in models that are not 
robust to even modest distributional shifts~\cite{DBLP:journals/corr/abs-1808-03305, 
	DBLP:journals/corr/abs-1711-11561, 
	DBLP:conf/nips/GeirhosTRSBW18,
	DBLP:conf/iclr/HendrycksD19}.
This is concerning because the non-static nature
of the real world may well cause
the distribution of inputs to shift during deployment,
and because
it is difficult for any finite training set 
to capture the full range of inputs that
may be encountered.
A model's lack of robustness likely
occurs due in part to over-reliance on non-robust features
that correlate well under the i.i.d.~assumption but 
stop providing useful information after a shift 
\cite{DBLP:conf/nips/IlyasSTETM19}.
Before deploying models,
especially in safety-critical contexts,
we must evaluate
their robustness to possible unknown shifts in the
distribution of encountered inputs.

In this work, we introduce a new method
that evaluates the robustness of neural
networks to changes that
(a) are context-sensitive perturbations
to features that vary in the training data and
(b) vary in granularity as we choose.
By \backtick context-sensitive', we mean
changes that are specific to the semantics
of the local objects in the image,
such as
object shape, position, size,
pose, colour and texture.
By \backtick granularity', we mean the
scope of a change in the image: a coarse-grained
change affects a large region of the image,
while the finest-grained change possible is to
a single pixel.

Why are these two properties desirable?
We ultimately want to evaluate robustness
to changes that might plausibly be encountered
at deployment.
For a given starting image, most
$\ell_p$ norm-constrained pixel perturbations
result in an image very unlikely to be encountered
at deployment; the few that are plausible are those
that represent a context-sensitive change to features
in a way that has been seen in the training data.
For instance, a perturbation that lengthens a table
by darkening a contiguous region of the image is much
more interesting than almost all perturbations
of a similar size to the same pixels.
Furthermore, by evaluating robustness to
such changes at different granularities,
we can investigate whether robustness in one
respect improves or worsens robustness in
other respects.

We obtain context-sensitive changes to features by
taking a pre-trained generative neural network
and perturbing its latent activation values
as it generates images. This works because
the neurons at different layers of a generator
encode the useful features
for generating images \cite{Bau:GDV:2019}.
We can control the granularity of the downstream
change by selecting the layers at which the
activations are perturbed: perturbations to
earlier layers
result in coarser-grained changes
(e.g., the shape of a building), while
later perturbations result in finer-grained
changes (e.g., the texture of a brick).

We use this method to evaluate
the robustness of state-of-the-art ImageNet
classifiers to context-specific feature
perturbations of different granularities.
Besides finding that these
models are not robust in this sense,
we make the surprising finding that
using classifiers adversarially that
were trained to be robust to
bounded pixel perturbations actually
\emph{decreases} robustness to coarse-grained
perturbations.
This may be because such classifiers must
necessarily depend more on coarser-grained features
of images than classifiers optimised for
accuracy on i.i.d.~inputs,
which tend to rely on fine-grained features
such as texture \cite{Geirhos:ICA:2019}.
Our results strengthen and expand upon
related findings from
\citet{DBLP:conf/nips/YinLSCG19},
who find that classifiers robust
to pixel-level perturbations are less robust
to corruptions of certain
context-insensitive features
such as artificial `fog'
and 2D sinusoids.

\section{Background}
\label{sec:background}

\paragraph{Robustness under distributional shift}
When training a discriminative neural network, the goal
is typically to minimise the expected loss
$\mathbb{E}_{(x, y) \sim P_0}[l(x, y;\; \theta)]$ with
respect to the model parameters $\theta$,
where $P_0$ is the training distribution 
over feature space $\sX$
and labels in $\sY$.
In real-world scenarios, however,
we cannot depend on the 
distribution at deployment remaining identical
to $P_0$ 
(i.e.~we cannot rely on the i.i.d.~assumption).
To ensure that models behave well in practice,
it is necessary to make distributionally
robust models: they should perform well even after 
a shift to their input distribution.
Studies have shown that 
there are many possible shifts to 
which classifiers are not robust, motivating 
a large body of literature which deals 
with identifying and 
correcting these problems \citep{
DBLP:journals/corr/abs-1808-03305, 
DBLP:journals/corr/abs-1711-11561, 
DBLP:conf/nips/GeirhosTRSBW18,
DBLP:conf/iclr/HendrycksD19, 
DBLP:journals/corr/AmodeiOSCSM16,
DBLP:conf/iclr/SinhaND18, 
DBLP:journals/corr/abs-1906-02899}.

\paragraph{Generative Adversarial Networks}
GANs are an approach to training generative
neural networks that map from a known
standard probability distribution
to the distribution of the training data.
See a tutorial for details \cite{Goodfellow:N2T:2017}. 
While other types of generative networks exist,
we focus on the use of GANs in this work for their
crispness. Generative
networks have been found to display an 
interesting property: different layers, and even
different neurons, encode different kinds of features
of the image.
Earlier layers tend to encode higher-level information about 
objects in the image, whereas later layers
deal more with ``low-level materials, edges, and colours''
\cite[p.7]{Bau:GDV:2019}.
In addition, it is possible to vary features
such as zoom, object position and rotation, simply
by moving the input to the model in a linear walk
\cite{DBLP:journals/corr/abs-1907-07171}.

\section{Related Work}
\label{sec:novelty}

\paragraph{Measuring robustness}
This paper builds on a body of work examining
trained models' robustness to a range of changes.
One approach is to apply a hand-selected range of possible
corruptions such as Gaussian noise or simulated
effects such as fog or motion blur.
Such robustness benchmarks have been created for datasets including
traffic signs~\cite{Temel:CCU:2017},
ImageNet \cite{DBLP:conf/iclr/HendrycksD19},
MNIST \cite{DBLP:journals/corr/abs-1906-02337}
and Cityscapes~\cite{Michaelis:BRI:2019}.
\citet{Snoek:CYT:2019} evaluate the robustness of
the calibration of classifiers' \emph{confidences}
to rotated and translated images, as well
as to out-of-distribution inputs
such as not-MNIST \cite{Bulatov:MLE:2011}.
Another approach is to gather new data,
either replicating original dataset creation processes
\cite{DBLP:conf/icml/RechtRSS19}
or deliberately gathering data representing a
challenging shift in
distribution \cite{DBLP:journals/corr/abs-1907-07174};
in both cases, classifiers were found to fail to generalise
to the new distributions.
Our paper builds on the foundations laid
by these works, providing automaticity by evaluating
robustness to \emph{learnt} features
that are sensitive to the local semantics.

\paragraph{Adversarial robustness}
There has been much recent work on
`adversarial examples': inputs
deliberately crafted by an adversary to fool
a model \cite{Gilmer:MTR:2018}.
In this literature, an attacker is modelled
as having certain capabilities to construct
pathological inputs, while the 
`defender' aims to create systems that are
robustly correct to all inputs within this
threat model.
Such attacks can be viewed as the
worst-case distributional shift within the threat
model.
The customary threat model is constrained
perturbations to the pixel values of a given
image \cite{Goodfellow:EAH:2015}, for which
adversarial robustness does not imply robustness
to more meaningful changes that induce large changes in
pixel values.
But there is a burgeoning interest in
new threat models that allow
\emph{meaningful} changes to given images.
While the purpose of our method is safety
(to evaluate models'
distributional robustness to plausible shifts in
the distribution of inputs),
rather than to serve as an
adversarial attack for security evaluations,
some existing `semantic attacks'
are related to our method.

\begin{figure*}[h]
	\centering
	\resizebox{\linewidth}{!}{
		\Large
		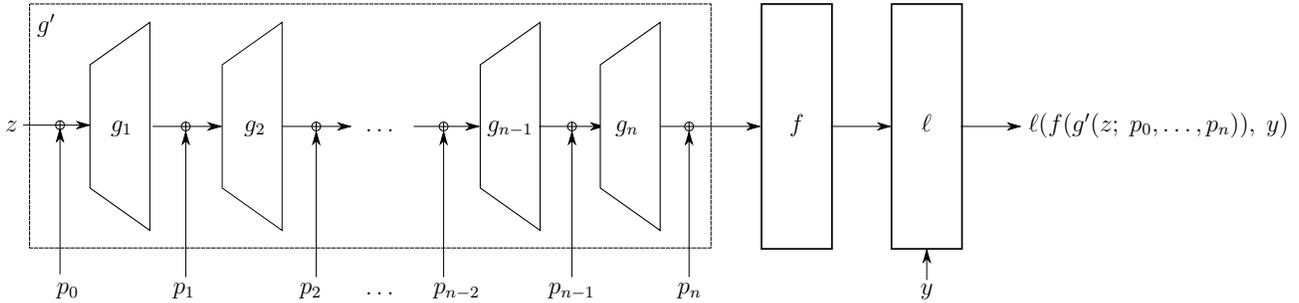}
	\caption{Illustration of a forward pass with
		perturbations to the latent activation values
		at $n$ layers in the generator network.}
	\label{fig:diagrams}
\end{figure*}

\paragraph{Semantic adversarial robustness}
Initially, semantic adversarial examples were
constructed using context-insensitive
hand-coded methods that
perturbed features such as colouring \cite{Hosseini:SAE:2018},
rotations and translations \cite{Engstrom:ETL:2019},
and corruptions such as blurring and
fog \cite{DBLP:conf/iclr/HendrycksD19} in an ad-hoc manner.
Another possible approach, albeit prohibitively expensive
for most domains, is to write an
invertible differentiable renderer and perturb
its parameters to effect semantic changes in
the scene \cite{DBLP:conf/iclr/LiuTLNJ19, DBLP:journals/corr/abs-1910-00727}.
More recently, context-sensitive methods have been proposed
that use generative models to avoid the need
to hand-code specific features.

\citet{Qiu:SGA:2019}
use a dataset labelled with various semantic features
to train a generative model that allows inputs
determining these features to be adversarially adjusted.
\citet{Bhattad:BBI:2019} utilise learnt colourisation
and texture-transfer models to identify worst-case
changes to the colour and texture of given images.
\citet{DBLP:journals/corr/abs-1912-03192}
adversarially compose
disentangled learned representations of different inputs.
Defense-GAN \cite{DBLP:conf/iclr/SamangoueiKC18} is not
an evaluation or attack, but an attempt to mitigate
attacks by projecting onto the GAN's learnt manifold.
\citet{Dunn:AGO:2019} train a GAN to output a
distribution of images that fools the target classifier.
Three imaginative papers construct semantic adversarial
examples using some search procedure to identify
a suitable input to a trained generative model
\cite{Zhao:GNA:2017, Song:CUA:2018, Wang:GSA:2020,Alzantot:GNL:2018};
\citet{DBLP:journals/corr/abs-2007-08450}
do likewise for a generator trained on
already-perturbed data.
We build on the common theme:
using a generative model to learn
meaningful context-specific semantic features.
However, our method is the first to perturb
the generator network's latent
activations---not just the input.
This leverages the \emph{full} range of
granularities of
feature representations learnt from the training dataset,
rather than just the maximally-coarse
semantics encoded in the input,
allowing for a much richer space of manipulations.
Evidence of this richness is given in our results.

\begin{figure*}
	\newlength{\mypicwidth}
	\setlength{\mypicwidth}{.18\textwidth}
	
	\centering
	\begin{subfigure}[b]{\mypicwidth}
		\centering
		\frame{\includegraphics[width=0.95\linewidth]{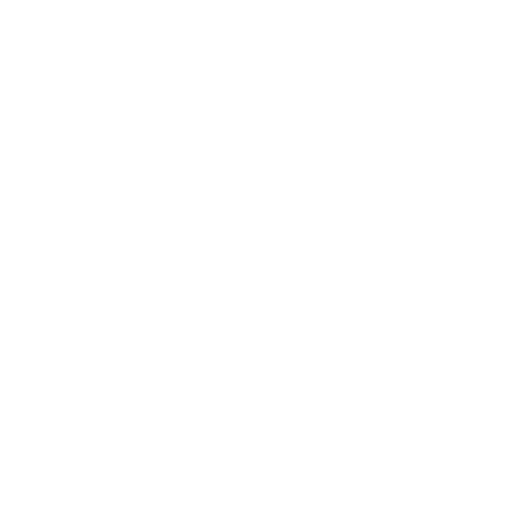}}
	\end{subfigure}\hfill%
	\begin{subfigure}[b]{\mypicwidth}
		\centering
		\frame{\includegraphics[width=0.95\linewidth]{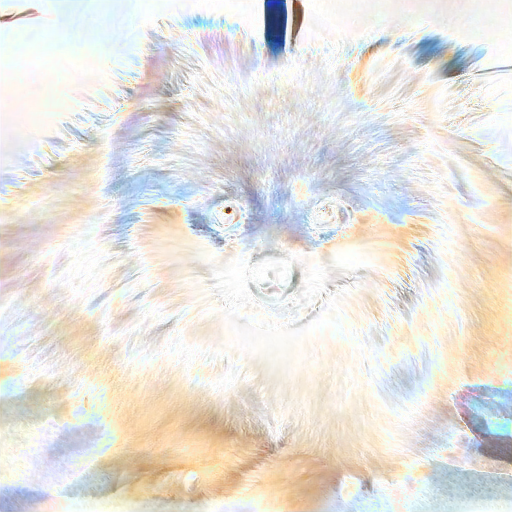}}
	\end{subfigure}\hfill%
	\begin{subfigure}[b]{\mypicwidth}
		\centering
		($\times 10$ for visibility) \\
		\frame{\includegraphics[width=0.95\linewidth]{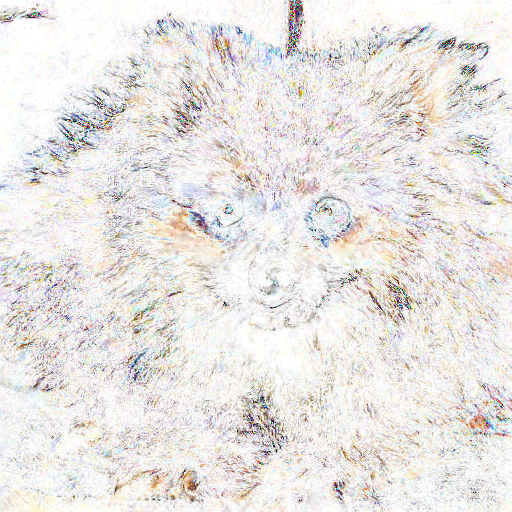}}
	\end{subfigure}\hfill%
	\begin{subfigure}[b]{\mypicwidth}
		\centering
		($\times 25$ for visibility) \\
		\frame{\includegraphics[width=0.95\linewidth]{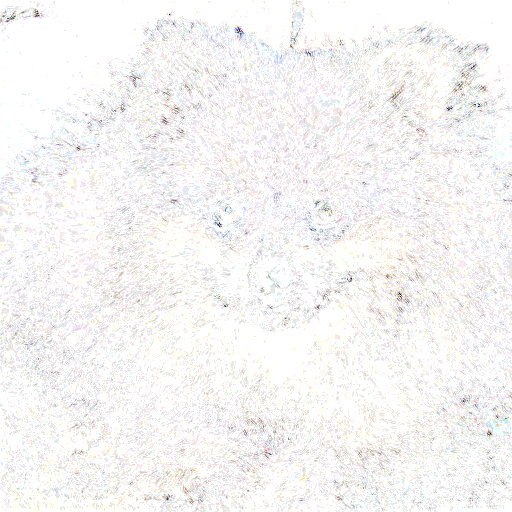}}
	\end{subfigure}\hfill%
	\begin{subfigure}[b]{\mypicwidth}
		\centering
		($\times 5$ for visibility) \\
		\frame{\includegraphics[width=0.95\linewidth]{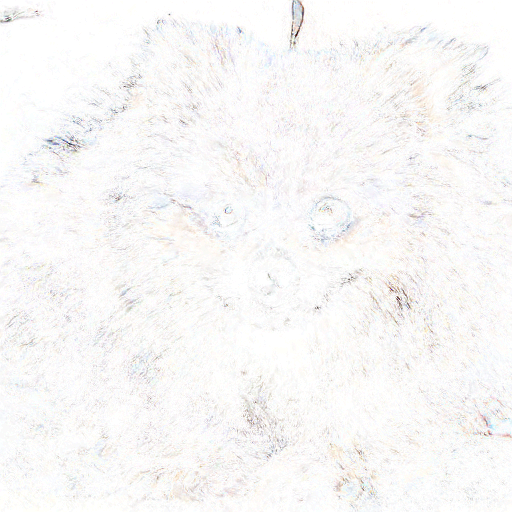}}
	\end{subfigure}
	
	\vspace{0.2cm}
	
	\begin{subfigure}[t]{\mypicwidth}
		\centering
		\frame{\includegraphics[width=0.95\linewidth]{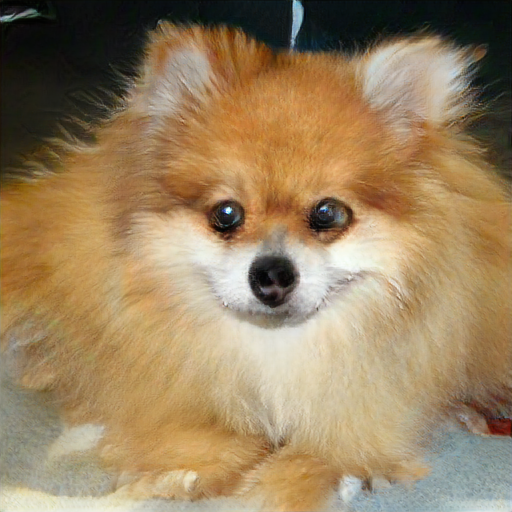}}
		\caption{Unperturbed.
		}
		\label{fig:depths-unpert3}
	\end{subfigure}\hfill%
	\begin{subfigure}[t]{\mypicwidth}
		\centering
		\frame{\includegraphics[width=0.95\linewidth]{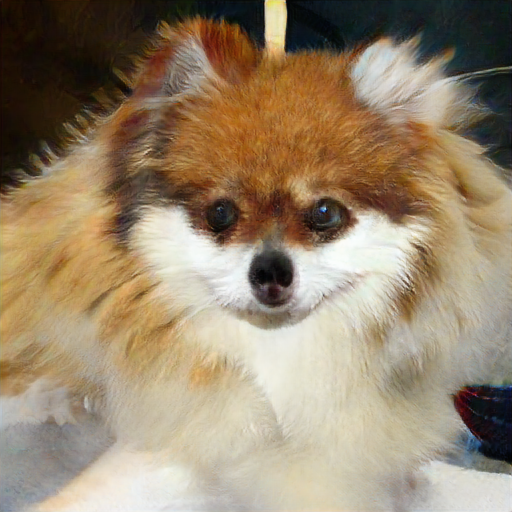}}
		\caption{First six layers.
		}
		\label{fig:depths-first3}
	\end{subfigure}\hfill%
	\begin{subfigure}[t]{\mypicwidth}
		\centering
		\frame{\includegraphics[width=0.95\linewidth]{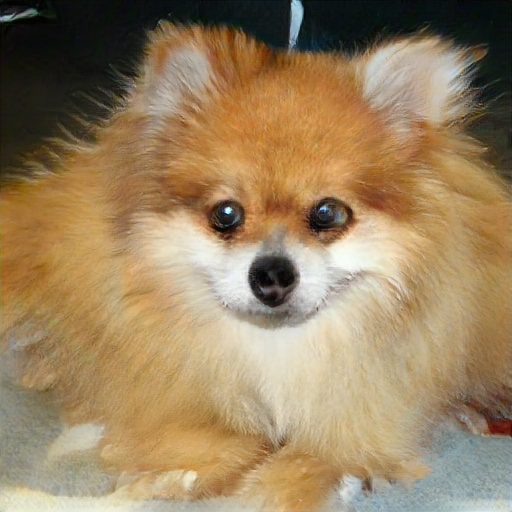}}
		\caption{Middle six layers.
		}
		\label{fig:depths-middle3}
	\end{subfigure}\hfill%
	\begin{subfigure}[t]{\mypicwidth}
		\centering
		\frame{\includegraphics[width=0.95\linewidth]{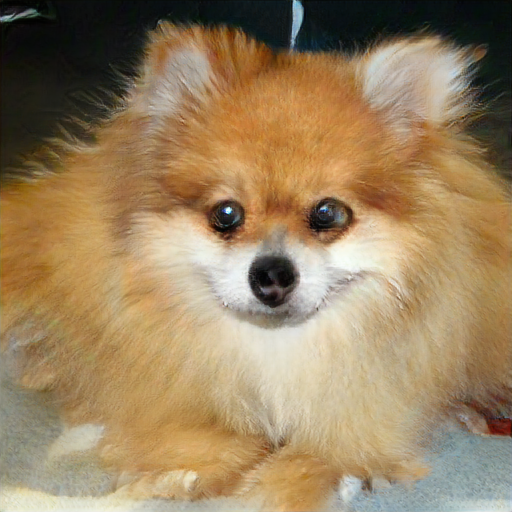}}
		\caption{Last six layers.
		}
		\label{fig:depths-last3}
	\end{subfigure}\hfill%
	\begin{subfigure}[t]{\mypicwidth}
		\centering
		\frame{\includegraphics[width=0.95\linewidth]{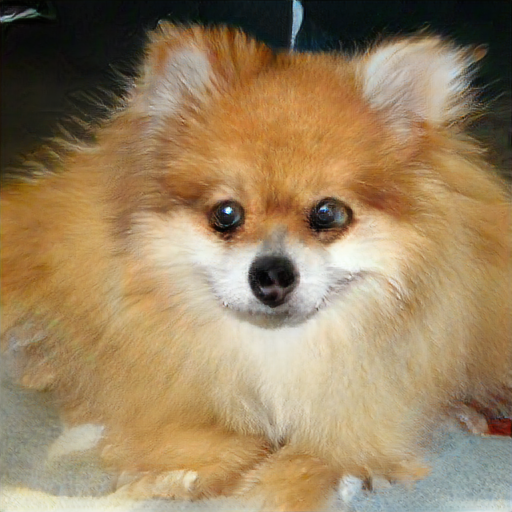}}
		\caption{All eighteen layers.
		}
		\label{fig:depths-all3}
	\end{subfigure}
	
	\vspace{0.4cm}
	
	\begin{subfigure}[b]{\mypicwidth}
		\centering
		\frame{\includegraphics[width=0.95\linewidth]{new_advex/volcs_depths/whitebox.png}}
	\end{subfigure}\hfill%
	\begin{subfigure}[b]{\mypicwidth}
		\centering
		\frame{\includegraphics[width=0.95\linewidth]{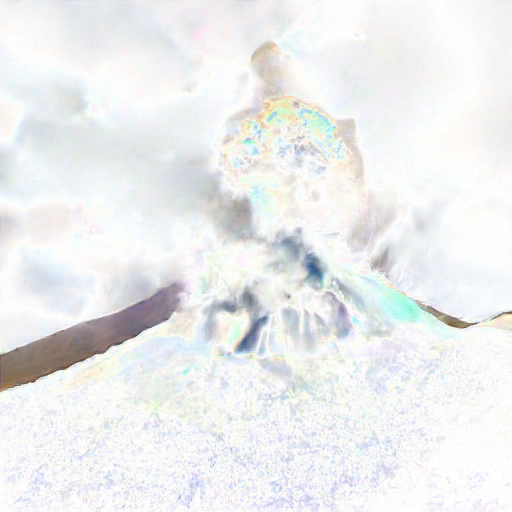}}
	\end{subfigure}\hfill%
	\begin{subfigure}[b]{\mypicwidth}
		\centering
		($\times 10$ for visibility) \\
		\frame{\includegraphics[width=0.95\linewidth]{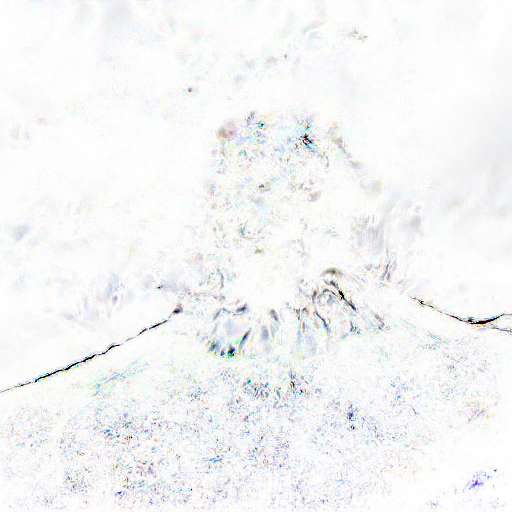}}
	\end{subfigure}\hfill%
	\begin{subfigure}[b]{\mypicwidth}
		\centering
		($\times 25$ for visibility) \\
		\frame{\includegraphics[width=0.95\linewidth]{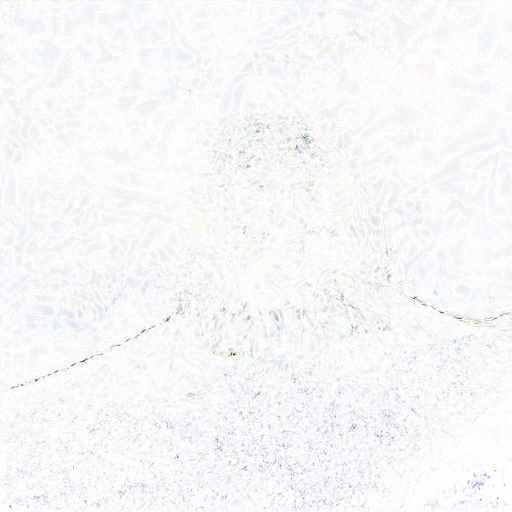}}
	\end{subfigure}\hfill%
	\begin{subfigure}[b]{\mypicwidth}
		\centering
		($\times 5$ for visibility)
		\frame{\includegraphics[width=0.95\linewidth]{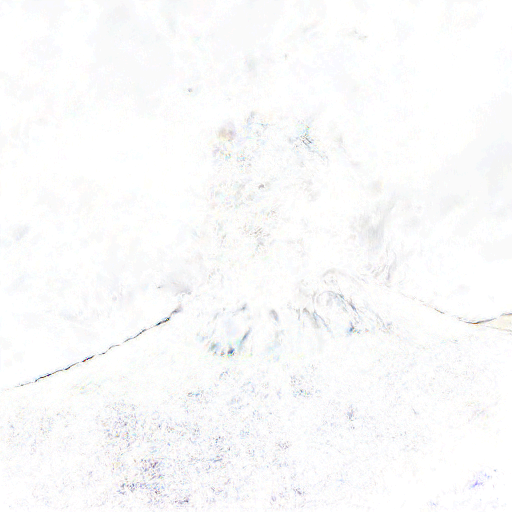}}
	\end{subfigure}
	
	\vspace{0.2cm}
	
	\begin{subfigure}[t]{\mypicwidth}
		\centering
		\frame{\includegraphics[width=0.95\linewidth]{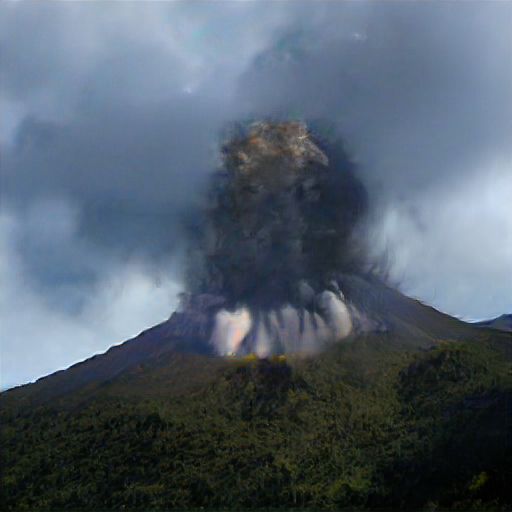}}
		\caption{Unperturbed.
		}
		\label{fig:depths-unpert2}
	\end{subfigure}\hfill%
	\begin{subfigure}[t]{\mypicwidth}
		\centering
		\frame{\includegraphics[width=0.95\linewidth]{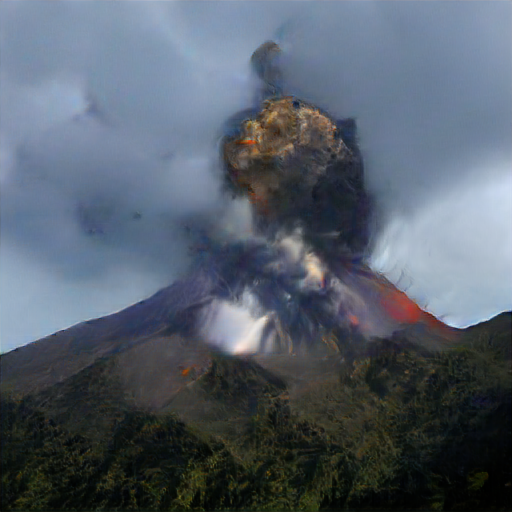}}
		\caption{First six layers.
		}
		\label{fig:depths-first2}
	\end{subfigure}\hfill%
	\begin{subfigure}[t]{\mypicwidth}
		\centering
		\frame{\includegraphics[width=0.95\linewidth]{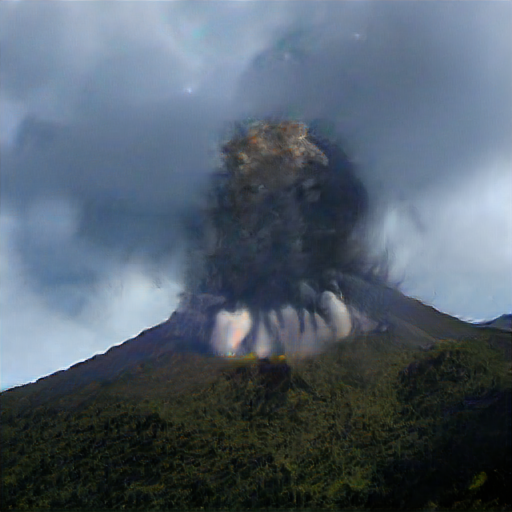}}
		\caption{Middle six layers.
		}
		\label{fig:depths-middle2}
	\end{subfigure}\hfill%
	\begin{subfigure}[t]{\mypicwidth}
		\centering
		\frame{\includegraphics[width=0.95\linewidth]{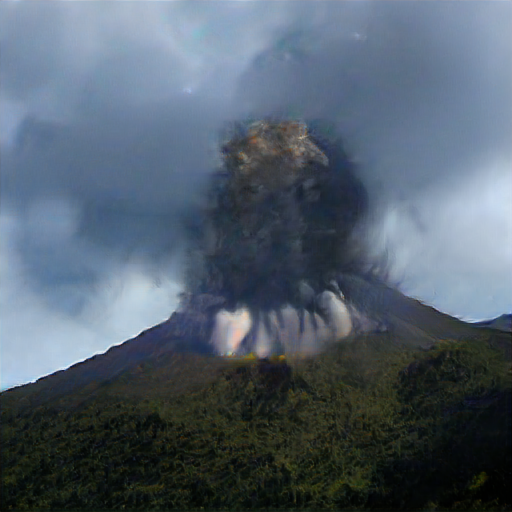}}
		\caption{Last six layers.
		}
		\label{fig:depths-last2}
	\end{subfigure}\hfill%
	\begin{subfigure}[t]{\mypicwidth}
		\centering
		\frame{\includegraphics[width=0.95\linewidth]{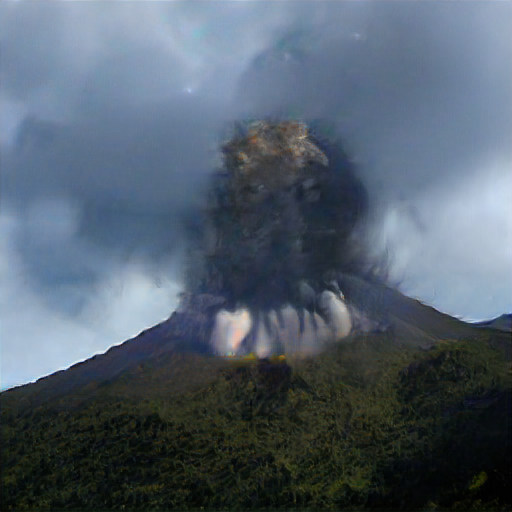}}
		\caption{All eighteen layers.
		}
		\label{fig:depths-all}
	\end{subfigure}
	
	\vspace{0.4cm}
	
	\begin{subfigure}[b]{\mypicwidth}
		\centering
		\frame{\includegraphics[width=0.95\linewidth]{new_advex/volcs_depths/whitebox.png}}
	\end{subfigure}\hfill%
	\begin{subfigure}[b]{\mypicwidth}
		\centering
		\frame{\includegraphics[width=0.95\linewidth]{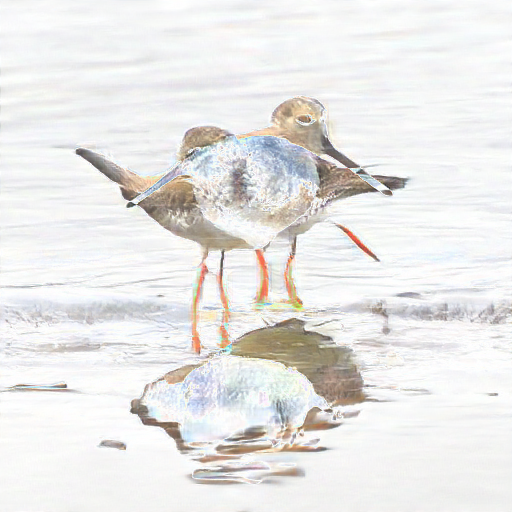}}
	\end{subfigure}\hfill%
	\begin{subfigure}[b]{\mypicwidth}
		\centering
		($\times 10$ for visibility) \\
		\frame{\includegraphics[width=0.95\linewidth]{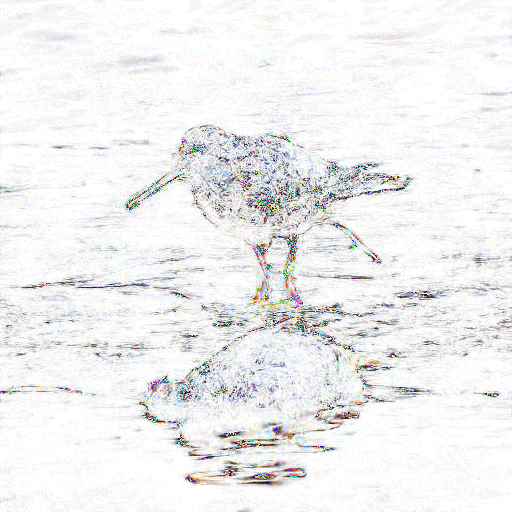}}
	\end{subfigure}\hfill%
	\begin{subfigure}[b]{\mypicwidth}
		\centering
		($\times 25$ for visibility) \\
		\frame{\includegraphics[width=0.95\linewidth]{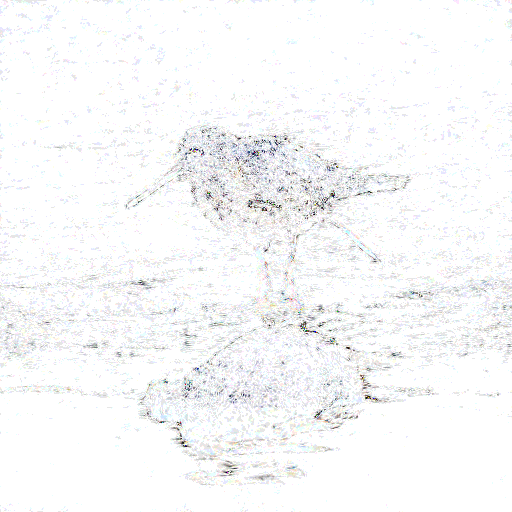}}
	\end{subfigure}\hfill%
	\begin{subfigure}[b]{\mypicwidth}
		\centering
		($\times 5$ for visibility) \\
		\frame{\includegraphics[width=0.95\linewidth]{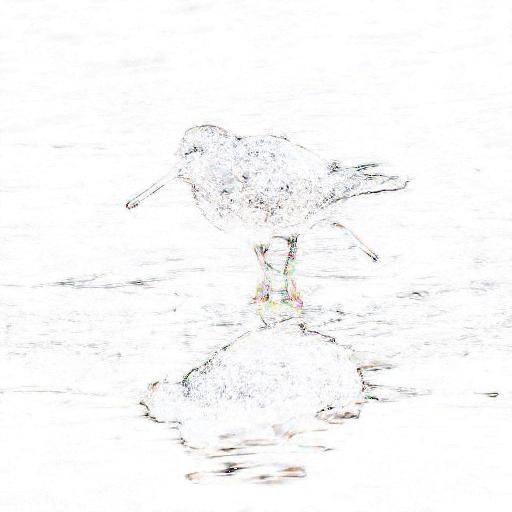}}
	\end{subfigure}
	
	\vspace{0.2cm}
	
	\begin{subfigure}[t]{\mypicwidth}
		\centering
		\frame{\includegraphics[width=0.95\linewidth]{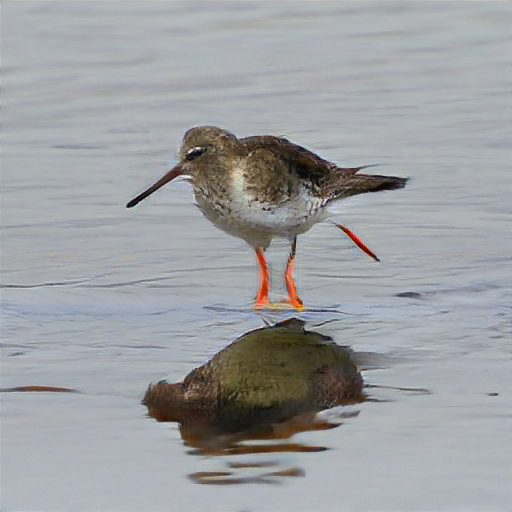}}
		\caption{Unperturbed.
		}
		\label{fig:depths-unpert}
	\end{subfigure}\hfill%
	\begin{subfigure}[t]{\mypicwidth}
		\centering
		\frame{\includegraphics[width=0.95\linewidth]{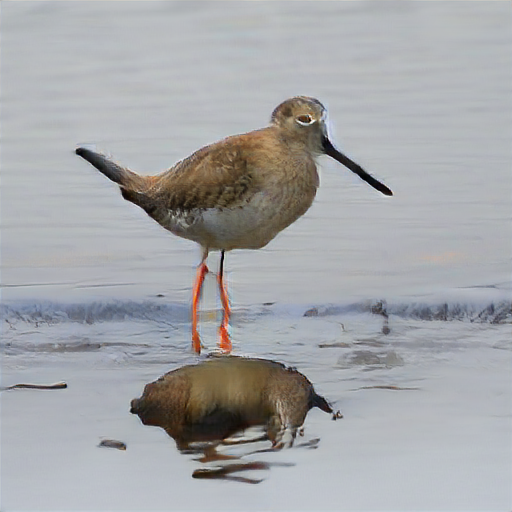}}
		\caption{First six layers.
		}
		\label{fig:depths-first}
	\end{subfigure}\hfill%
	\begin{subfigure}[t]{\mypicwidth}
		\centering
		\frame{\includegraphics[width=0.95\linewidth]{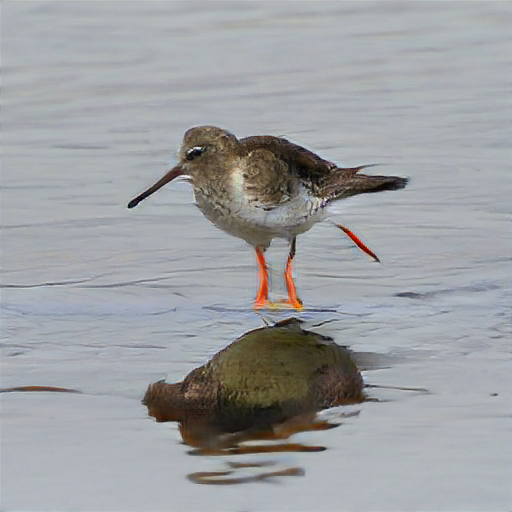}}
		\caption{Middle six layers.
		}
		\label{fig:depths-middle}
	\end{subfigure}\hfill%
	\begin{subfigure}[t]{\mypicwidth}
		\centering
		\frame{\includegraphics[width=0.95\linewidth]{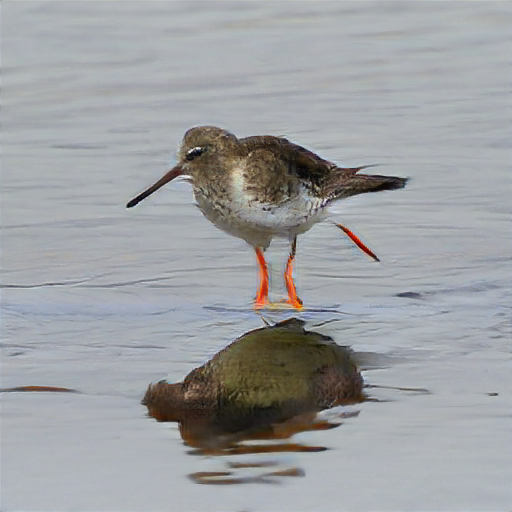}}
		\caption{Last six layers.
		}
		\label{fig:depths-last}
	\end{subfigure}\hfill%
	\begin{subfigure}[t]{\mypicwidth}
		\centering
		\frame{\includegraphics[width=0.95\linewidth]{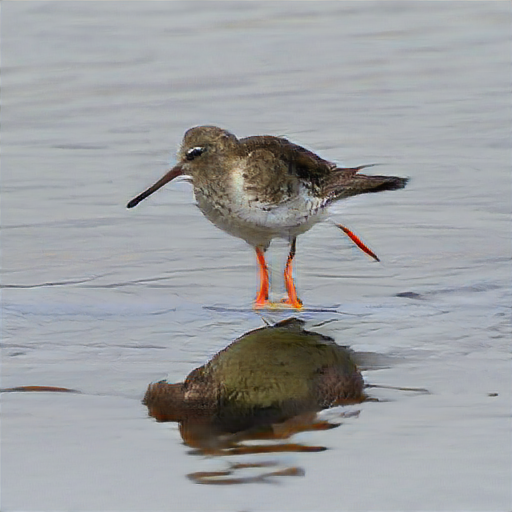}}
		\caption{All eighteen layers.
		}
		\label{fig:depths-all}
	\end{subfigure}
	\caption{Context-sensitive feature perturbations
		at different granularities, as controlled
		by perturbing activations at the
		generator layers indicated under each image.
		Differences with the unperturbed 
		image are shown above each perturbed image.
		The perturbed Pomeranians (dogs) are classified as \backtick red king crabs',
		the volcanos as \backtick goldfish',
		and redshanks (birds) as \backtick rams'.
	}
	\label{fig:depths}
\end{figure*}

\section{Method}
\label{sec:methods}

Suppose that we have a trained, differentiable
image classifier $f: \sX \to \R^{|\sY|}$
whose robustness we would like to evaluate,
where  $\sX = \R^{3 \times w \times h}$ is RGB pixel space and
$\sY$ is the set of class labels over which $f$ outputs a
confidence distribution.
Suppose that we also have a trained generator neural network
$g: \sZ \to \sX$, which maps from a standard
Gaussian distribution
over its input space $\sZ = \R^m$
to the distribution of training images.
Although we use the generator of a GAN,
a VAE or
any other generative model would be equally suitable.

Since a feedforward network is a sequence
of layers, we can consider $g$ to be a composition
of functions
$g = g_n \circ g_{n-1} \circ ... \circ g_1$.
For instance, in our main experiments, we decompose BigGAN \cite{Brock:LSG:2019} into
its residual blocks.
Here, $g_i: \sA_{i-1} \to \sA_i$ is the $i$th layer,
taking activations $a_{i-1} \in \sA_{i-1}$
from the previous layer and outputting the
resulting activation tensor $g_i(a_{i-1}) \in \sA_{i}$.
Splitting in this way
allow us to introduce a perturbation $p_i \in \sA_i$
to layer $i$'s activations, before continuing
the forward pass through the rest of the generator.
Given such a perturbation tensor
$p_i \in \sA_i$ for each layer $i$,
we can define perturbed layer functions
$g'_i(a_{i-1}) = g_i(a_{i-1}) + p_i$.
By performing such perturbations at every activation
space, we obtain the perturbed output of
the entire generator,
$g'(z ;\; p_0,\: ...,\: p_n) =
(g'_n \circ g'_{n-1} \circ ... \circ g'_1)(z+p_0)$.

Suppose that we have sampled some generator input $z$,
and automatically determined
the correct label~$y$ of its image under the generator
when unperturbed, $g'(z;\; 0,\: ...,\:0)$.
This is best done using a conditional
generator that also takes a label as input \cite{Mirza:CGA:2014},
but can also be achieved by assuming
that the classifier's initial output is correct.
We now need a procedure to identify suitable perturbation tensors
$\vp^* = (p_0^*, ..., p_n^*) \in \sA_0 \times \ldots \times \sA_n$
such that the classifier's
output on $g'(z;\; \vp^*)$ is not $y$ but some other label,
while the true label of $g'(z;\; \vp^*)$ remains $y$.

\paragraph{Ensuring misclassification}
We can define a loss function $\ell: \R^{|\sY|} \times \sY \to \R$
such that
$\ell(f(g'(z;\; \vp)),\; y)$ is minimised when
the classifier predicts any label but the correct $y$,
or such that $\ell(f(g'(z;\; \vp)),\; t)$ is
minimised when the classifier incorrectly predicts
target label $t$.
There are many possibilities, but in this paper
we focus on the latter case only,
using
$\ell(f(x),\; t) = \max_{j\in\sY}{f(x)_j}-f(x)_t$,
the variant found to be most effective
by \citet{Carlini:TET:2017}.
Noting that $\ell$, $f$ and each $g'_i$ are differentiable, we
use the usual backpropagation algorithm to compute
the derivative of $\ell(f(g'(z;\; \vp)),\; y)$ with respect
to $\vp$. We then use a gradient-descent
optimisation over $\vp$
to find a perturbation $\vp^*$ that minimises $\ell$.
By definition of $\ell$, this optimal $\vp^*$ will ensure that
the classifier mislabels perturbed image $g'(z;\; \vp^*)$.

\paragraph{Ensuring the true label remains unchanged}
But we also need the true label of $g'(z;\; \vp^*)$ to remain $y$,
else $f$ might in fact be predicting the correct label.
Our approach is to assume that \emph{small} changes
to an image will not change its correct label
(we verify this in our experiments).
So we would like to identify the smallest perturbation
that induces the kind of mislabelling we are investigating.
The approach we take is to constrain the maximum magnitude of the
perturbation---computed as the Euclidean norm of the
vector obtained by `flattening' and concatenating
the perturbation tensors $p_i$---and
gradually relaxing this constraint during optimisation until a
perturbation $\vp^*$ inducing the desired mislabelling is found.
The quicker the constraint is relaxed, the quicker a mislabelling
is found, but the larger the expected
perturbation.

\paragraph{Per-neuron perturbation scaling to promote uniformity}
The typical activation values of separate
neurons differ in scale, even within a single layer.
If one varies from $-1$ to $1$, while another
varies from $-0.1$ to $0.1$, then
a perturbation of magnitude $0.5$ is likely
to have quite different downstream effects
on the image depending on which of these neurons it
affects.
To correct for this, we scale the perturbation
for each neuron to the empirically-measured range.
That is, rather than adding perturbation tensor
$p_i$ directly to the activation tensor at layer $i$,
we add $p_i \odot \sigma_i$, where $\odot$ is
element-wise multiplication and $\sigma_i$ is
an empirically-measured tensor of standard deviations
of the activation values at layer $i$.

In future work, it may be desirable to further
fine-tune the scale of the perturbations
applied to each neuron. In the present work,
however, the above scaling procedure is
sufficient to normalise the downstream effect
of perturbing different neurons; if it were
insufficient, then too many of the perturbed
images would no longer be recognisable as their
original class.

\section{Experimental Evaluation}
\label{sec:results}

In this section, we apply our method to evaluate
the robustness of state-of-the-art classifiers to
context-specific feature perturbations
of different granularity.
Primarily, we evaluate ImageNet-1K classification \cite{Russakovsky:ILS:2015},
perturbing the features learnt by a
pre-trained BigGAN \cite{Brock:LSG:2019}.
We evaluate two standard classifiers, and two 
`robust' classifiers adversarially trained against 
bounded pixel-space perturbations.
First, the state-of-the-art
on ImageNet, EfficientNet-B4 with
NoisyStudent training \cite{DBLP:journals/corr/abs-1911-04252}.
This was the highest-accuracy classifier for which
pre-trained weights were available.
Next, the standard ResNet50 classifier
\cite{DBLP:conf/cvpr/HeZRS16}.
Finally, two ResNet50 classifiers adversarially trained against 
pixel-space perturbations: one from \citet{robustness}
trained using an $\ell_2$-norm PGD attack with
radius $\eps = 0.3$, and another
from \citet{DBLP:conf/iclr/WongRK20},
trained with the FGSM attack
for robustness against
$\ell_{\infty}$ with $\eps = 4/255$. 
See Appendix~\ref{app:models} for further details.

We focus
exclusively on
targeted misclassifications, for which
a randomly predetermined target label $t \in \sY$ is
chosen for each input, since
failure in this case implies significant weakness.
A classifier is more robust to a class of
perturbations if larger-magnitude perturbations
of that kind are required to induce the 
targeted classification.
Recall that we gradually relax the constraint
on the magnitude of the perturbations to activation values:
by measuring the smallest magnitude for which the
classifier outputs the target class $t$, we can build
a picture of how robust the model is to the perturbations.

We cannot guarantee that the images originally produced by 
BigGAN would be labelled by humans as their intended class
(GANs are imperfect), or that the perturbations do not 
change the class of the image. The risk in making 
visible and varied perturbations is that it becomes 
difficult to ensure these not change the class. 
To eliminate these risks, 
we use five independent human judges to 
vote on the class of the images.
As in the original ImageNet labelling
protocol, a majority is used to decide the label.
It is also possible to avoid human labelling, by 
fixing a maximum perturbation magnitude---although
this does introduce a trade-off between false
positives and negatives.
See Appendix~\ref{app:no-human} for details.

Our experiments proceed as follows: we randomly 
generate an unperturbed image of class $y$; we
skip this image if the classifier does not predict 
class $y$ to begin with. We randomly select 
a target $t$, and find a perturbation which
induces this misclassification. Human labellers 
vote on whether the unperturbed image is truly 
of class $y$; the image is skipped if not. 
If the image is not skipped, the labellers 
who voted to keep the unperturbed image vote on whether the 
perturbed image is (still) of class $y$. 

In this way, we obtain a set of correctly-classified
unperturbed images paired with incorrectly-classified 
perturbed counterparts for which we know both 
the perturbation magnitude, and
whether they were successful (maintained the class of the image).
We then evaluate robustness by looking at how quickly 
the number of successful perturbations grows as 
we allow for larger and larger perturbation magnitudes.
To evaluate robustness to changes to different
feature granularities,
we repeat this robustness analysis, but restrict
the perturbations to affect different
subsets of layers.
For full details of our experiments,
refer to Appendix~\ref{app:details}.

To demonstrate that our method readily
generalises, we also run our experiments
on a smaller scale
on two further datasets.
First, the much simpler
(and correspondingly easier to robustly classify)
MNIST dataset \cite{LeCun:MHD:}, using one classifier
optimised for i.i.d.~accuracy, and one adversarially
trained to be robust to an $\ell_2$-norm projected gradient descent
attack with $\eps = 0.3$.
Then, another high-resolution dataset:
CelebA-HQ,
with our method applied to a Progressive GAN
\cite{DBLP:conf/iclr/KarrasALL18}.

\subsection{Results}

\begin{figure*}[!ht]
	\centering
	\newlength{\graphwidth}
	\setlength{\graphwidth}{.42\textwidth}
	\begin{subfigure}[t]{\graphwidth}
		\centering
		\includegraphics[width=\linewidth]{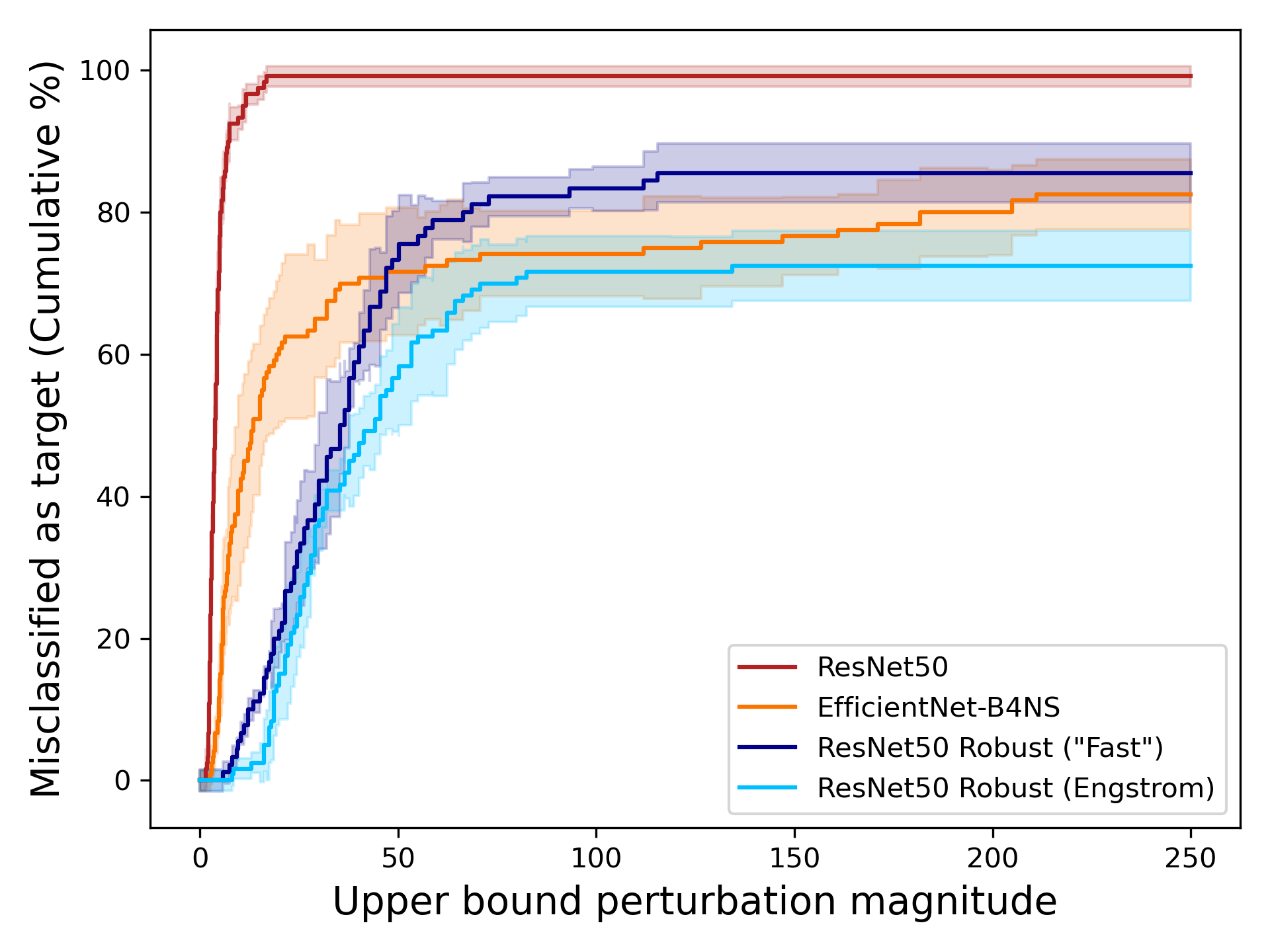}
		\caption{Activation values perturbed at all BigGAN layers.}
		\label{fig:All_layers}
	\end{subfigure}\hfill%
	\begin{subfigure}[t]{\graphwidth}
		\centering
		\includegraphics[width=\linewidth]{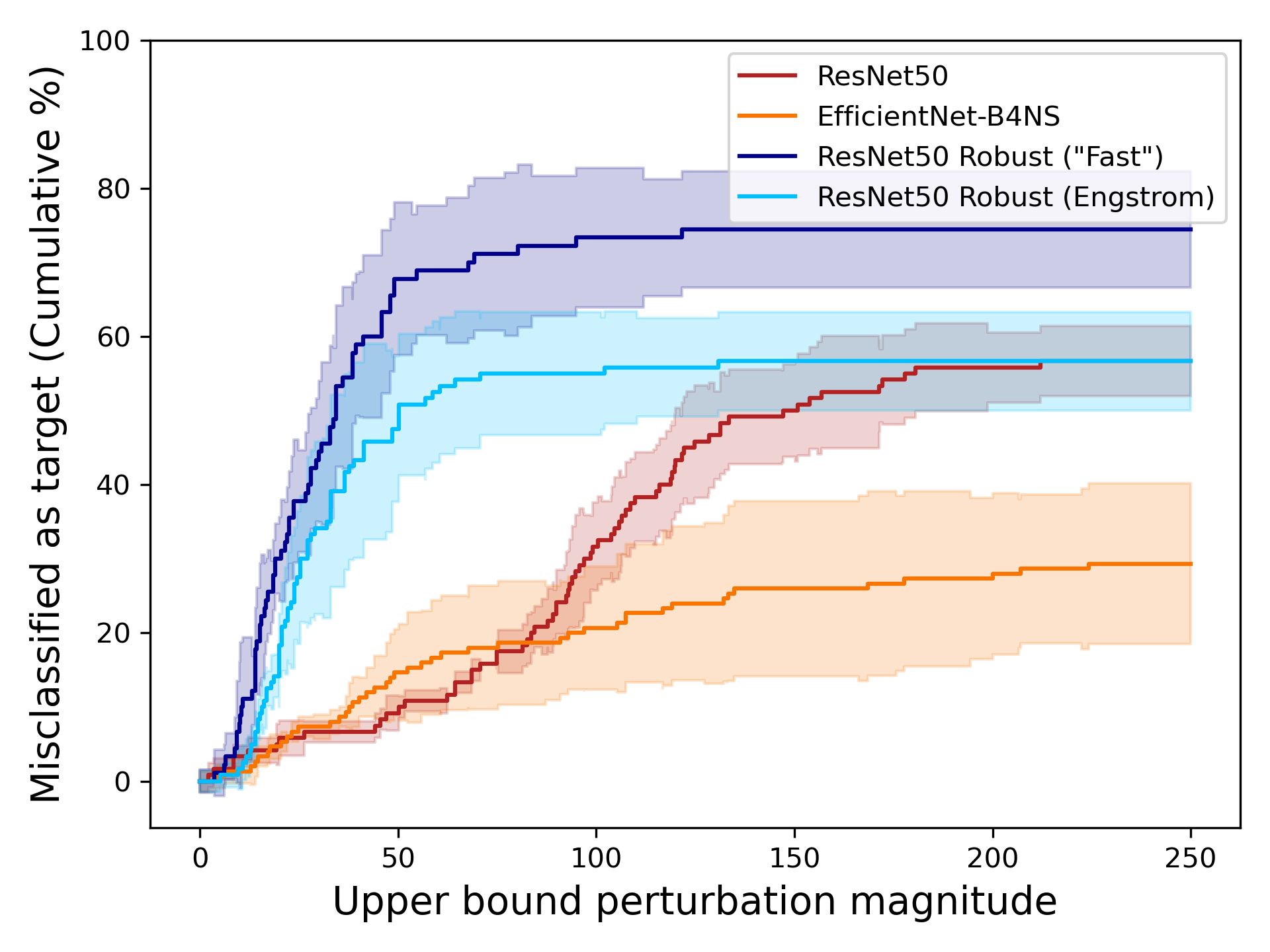}
		\caption{Activation values perturbed in the first six layers only.}
		\label{fig:1/3}
	\end{subfigure}
	
	\vspace{3mm}
	
	\begin{subfigure}[t]{\graphwidth}
		\centering
		\includegraphics[width=\linewidth]{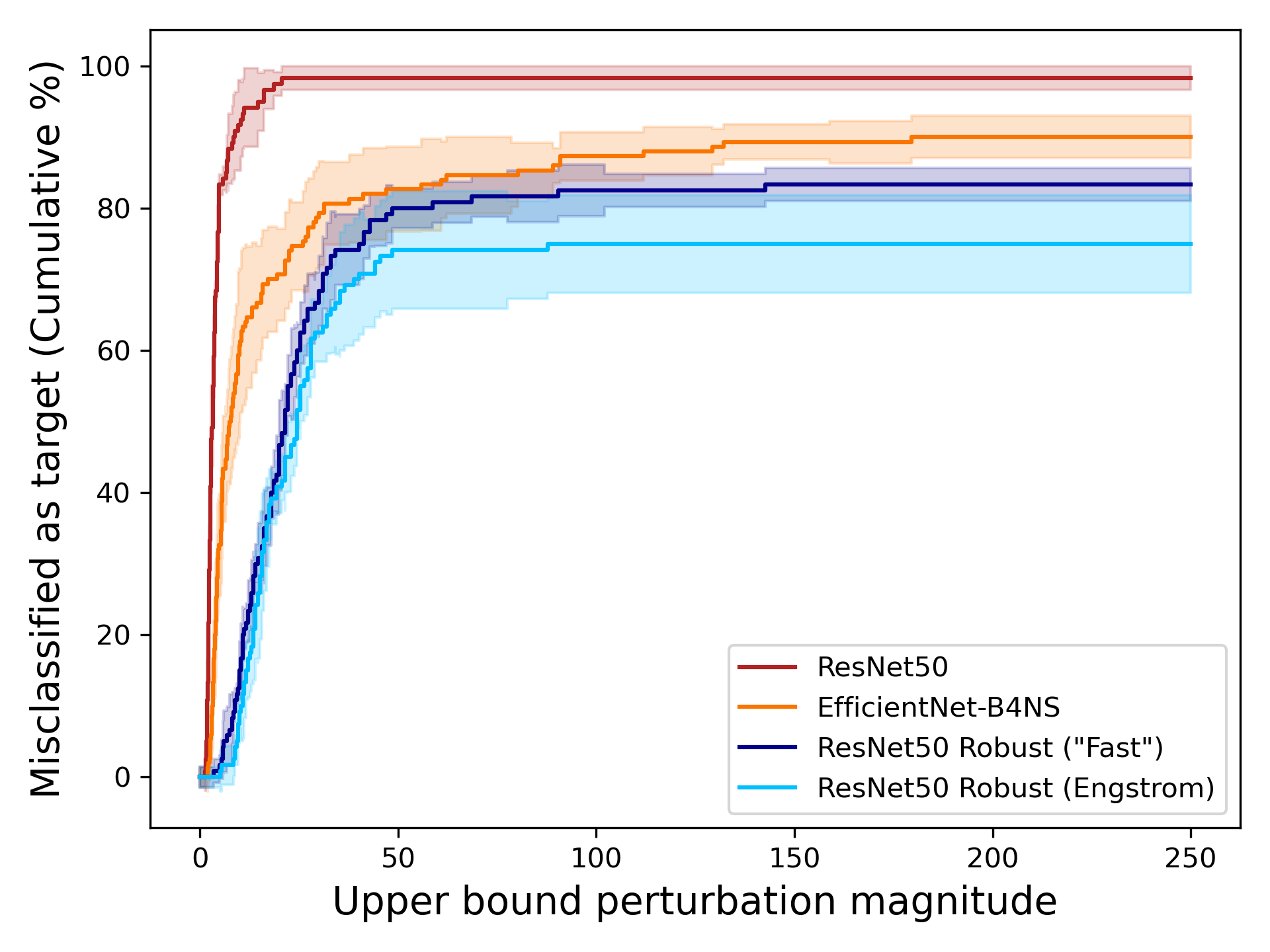}
		\caption{Activations perturbed in the middle six layers only.}
		\label{fig:2/3}
	\end{subfigure}\hfill%
	\begin{subfigure}[t]{\graphwidth}
		\centering
		\includegraphics[width=\linewidth]{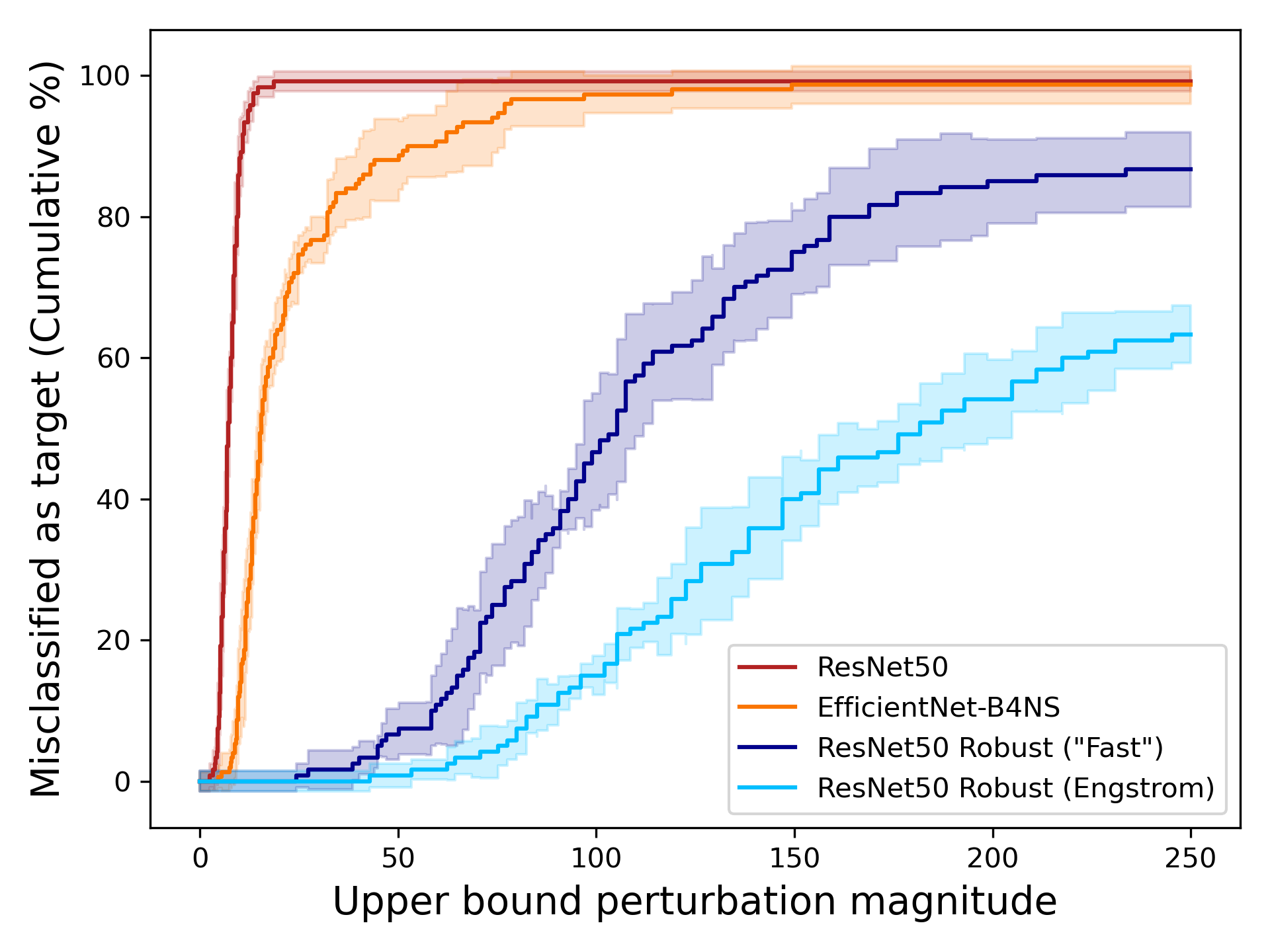}
		\caption{Activation values perturbed in the last six layers only.}
		\label{fig:3/3}
	\end{subfigure}\hfill
	
	\caption{
		Graphs showing how the cumulative proportion
		of perturbations
		that induce the targeted misclassification
		increases with maximum perturbation magnitude.
		The steeper the line, the less robust the
		classifier to that perturbation type.
		The lines and translucent areas
		shown are the means and
		standard deviations between several experiments
		of 30 images each.
	}
	\label{fig:graphs}
\end{figure*}

Table~\ref{tab:magnitudes} reports the average
magnitude of the misclassification-inducing
perturbations.
Figure~\ref{fig:graphs} elaborates on this,
plotting the relationships
between perturbation magnitude and the
cumulative proportion of inputs for which this magnitude
(or smaller)
is sufficient to cause the classifier to
predict the target label.
For each type of perturbation,
for each classifier,
$192 \pm 20$ (minimum $158$) unperturbed images were labelled by
the human judges,
of which $53 \pm 8$ (max $69$) images were rejected
by the majority for not matching the intended label.
Note that this latter quantity depends only on the
pre-trained generator, not our method.
See above and Appendix~\ref{app:prod_graphs}
for full details of our procedure.

Of course, even state-of-the-art GANs
generate images that are not photorealistic.
But we note that photorealism from the generative
model is not necessary for our method.
We want to trust our classifiers
to behave correctly on images that are
unambiguously of a certain class: all that is
necessary is that the generated images have
this property. Our standard for labelling
is a majority vote among our five judges;
all images that meet this criterion yet
are misclassified by a model are weaknesses
of the model.

\subsection{Discussion of Results}
\label{sec:discussion}

\paragraph{Qualitative results}
The results in Figure~\ref{fig:depths}
and Appendix~\ref{app:imagenet-examples}
demonstrate a range of the context-specific
feature perturbations
that our method produces,
and shows that perturbations at different
layers produce downstream changes of different
granularities.
We are publishing the full dataset of perturbed
images used in our experiments; see
Appendix~\ref{app:imagenet-examples}.

\paragraph{None of the classifiers are robust to any granularity of perturbation}
Figure~\ref{fig:All_layers} shows that for all four
classifiers, our method finds
misclassification-inducing perturbations
for over 80\% of the initial generated images, 
even with relatively small perturbation magnitudes.
This result is consistent with the notion that
trained classifiers have learnt to rely on
spurious (or at least fragile)
feature correlations that may not generalise
beyond the training regime.
For instance, relying on background colour
to identify the foreground object may
work well if this correlation holds---as it would both
during normal i.i.d.~training and adversarial training---but
this should not be relied upon at deployment.

\begin{table}
	\centering
	\caption{Mean magnitudes of misclassification-inducing perturbations,
	for different classifiers (rows) and different
	layers in the generator
	at which activations are perturbed (columns).
	Compare results within each column to compare
	robustness to the same granularity of perturbation.
	}
	\label{tab:magnitudes}
	\begin{tabularx}{\linewidth}{lcccc}
		\toprule
		&  All layers &  First 6  &  Middle 6  &   Last  6  \\
		\midrule
		Engstrom   & 36  & 33 & 21  & 141  \\
		``Fast''  & 35  & 29 & 22  & 102 \\
		EfficientNet   & 36   & 97 & 22   & 24 \\
		ResNet50      & 4.2 & 89 & 4.2 & 7.4 \\
		\bottomrule
	\end{tabularx}
\end{table}

\paragraph{Different generator layers encode meaningfully-different features}
Since the results show that the classifiers behave
significantly differently when perturbations are
restricted to different groups of layers,
the kinds of features being changed must meaningfully
differ. This is clear evidence that perturbing
intermediate activations offers a much richer feature
space than perturbing (say) the generator input only.

\paragraph{Pixel-space robustness improves robustness to finer-grained perturbations}
The lower average magnitudes required for the pixel-robust
classifiers when perturbing the
final six layers, as seen in the last column of
Table~\ref{tab:magnitudes}
and the correspondingly flatter curves in
Figure~\ref{fig:3/3}, demonstrate that adversarial training
generalises somewhat to confer robustness to
fine-grained feature perturbations.
The slightly gentler gradient at the beginning of
Figure~\ref{fig:2/3} suggests that this even provides some limited
robustness to small perturbations of medium granularity.
In both cases, this may be because the changes
fall within or nearby the pixel-space $\ell_p$-norm ball that
the classifier is trained to be robust within.

\paragraph{But robustness to pixel-space perturbations is a double-edged sword}
Perturbations to activations in the early layers of a generator
induce context-sensitive, coarse-grained changes to 
the features of an image.
These have a large magnitude when measured in pixel space,
so it is unsurprising that classifiers trained
to be robust to norm-constrained pixel perturbations
do not have improved robustness to such feature perturbations.
More surprisingly, Figure~\ref{fig:1/3} and
Table~\ref{tab:magnitudes} show that
pixel-space robustness in fact considerably \emph{worsens}
robustness to the coarse-grained features
encoded in the first six
generator layers.
This may be because non-robust standard classifiers ordinarily
depend mainly on fine-grained features
such as texture \cite{Geirhos:ICA:2019}.
Conversely, pixel-space robust classifiers
have been trained to ignore these fine-grained features, 
and so depend instead on coarse-grained features.
But they can still rely on
fragile correlations in the coarse-grained features, so
their robustness to context-sensitive coarse-grained feature
perturbations is decreased.

There is already some evidence that 
classifiers optimised for
robustness to constrained adversarial pixel
perturbations seem to have decreased robustness to
corruptions concentrated in the
low-frequency Fourier domain
\cite{DBLP:conf/nips/YinLSCG19},
decreased robustness to invariance-based
attacks that change the true label
but maintain the model's prediction
\cite{DBLP:journals/corr/abs-2002-04599}
and little robustness to
various context-insensitive corruptions not encountered
at training time \cite{Kang:TRA:2019a}.
Our finding that such models also have
significantly decreased robustness to
coarse-grained context-sensitive feature perturbations
strengthens and generalises these results.

\paragraph{MNIST}
The simplicity of the MNIST classification task
suggests that constructing a robust classifier
for MNIST should be significantly easier than for ImageNet.
We find that adversarial training against pixel perturbations
does not improve robustness to coarse-grained
perturbations on MNIST, but neither does it worsen it.
This is likely because the simplicity and
low resolution of the dataset
significantly reduces the range of
possible granularities, relative to ImageNet.
See Appendix~\ref{app:mnist} for results and discussion.

\paragraph{CelebA-HQ}
In Appendix~\ref{app:celeba}, we show that
our method also easily generalises to work for
a pre-trained Progressive GAN~\cite{DBLP:conf/iclr/KarrasALL18}
on the CelebA-HQ dataset.
This neatly demonstrates that our method is general,
in that it does not depend on properties of any dataset
or model.

Note that the results on ImageNet, CelebA-HQ, 
and MNIST show that the method described in
Section~\ref{sec:methods} is sufficient.
On all three we did \emph{no} hyper-parameter tuning.
We used the obvious choice for where to make perturbations, 
and did not tune size of the perturbations at each neuron.

\section{Conclusion}
Since the i.i.d.~assumption
cannot be relied upon during deployment, it is necessary
for our models to remain performant when
given inputs different from those in its
training distribution.
In this paper, we have introduced a new
method that evaluates robustness of image classifiers to
a rich class of such inputs:
by dynamically
perturbing the intermediate activation values
of trained generative neural networks
we produce context-sensitive perturbations
to meaningful learnt features of different
granularities.
This allows evaluation of robustness to changes
varying from
coarse-grained properties
such as object shape and colour
(encoded in earlier layers)
to fine-grained edges and textures (encoded in later layers).

Perhaps unsurprisingly, we
find that modern state-of-the-art ImageNet classifiers 
are not robust to context-sensitive features perturbations
at any granularity.
More surprisingly, while
classifiers optimised for robustness
to $\ell_p$ norm-bounded pixel perturbations are indeed more robust to 
fine-grained feature perturbations,
this is to the detriment of their 
robustness to coarse-grained feature perturbations.
Besides the obvious need for improved models,
our findings motivate the need for a deeper
understanding of robustness of different kinds,
and more comprehensive 
meaningful evaluations of models.
We hope that the present work is a step in the right direction.

\section*{Ethics Statement}

Our method produces context-dependent
feature perturbations of different granularities.
We hope that this line of work will eventually
produce tools that can be used to thoroughly
evaluate the robustness of machine learning systems.
This would be extremely useful: we should only
give such systems responsibility when we are
sure that we can rely on them to behave well
even if they are presented with inputs that
do not exactly match the kind they were trained
on---an inevitability in our ever-changing world.

But we should point out that although our method
is a very helpful step towards a thorough robustness
evaluation, it is currently insufficient.
GANs are known to \backtick drop modes', meaning
that not all kinds of variation present in the
training data are learnt. Moreover, it certainly
is not clear that all kinds of variation to which
we desire robustness can even in principle be
represented as perturbations to the intermediate
activations of a generator.
Therefore, our method must be viewed as one tool
in a box of evaluation methods, and as an invaluable
step towards more comprehensive evaluations; it must
not be relied on by itself.
We note also that even a thorough robustness evaluation
would be insufficient, since there may be other necessary
properties of models such as maintaining privacy
or fairness.

Generative modelling is well-known to be dual-use, in
the sense that generative models can be used for harmful
as well as beneficial purposes. For example, they
can be used to generate \backtick deepfake' videos
that can deceive the viewer into thinking something
untrue on an important subject \cite{DBLP:conf/avss/GueraD18}.
Our work is not an advancement in generative modelling.
Nor is it the first to identify that different
intermediate neurons control different meaningful
features of outputs \cite{Bau:GDV:2019}.
But this paper may draw more
attention to this true fact, which may be helpful
to malicious individuals creating harmful
\backtick deepfake' videos.
Given the importance of understanding
and mitigating robustness weaknesses in
deep learning models, we believe that the
benefits of our work easily outweigh the costs.

In this paper, we find that robustness in the sense
that most contemporary work focuses on---that is,
robustness to $\ell_p$ norm-constrained pixel-space
perturbations---is not only insufficient for robustness
against high-level meaningful changes of the kind that
might be encountered during deployment, but in fact
\emph{worsens} such robustness.
It is important that the research community and
especially organisations deploying commercial
deep learning applications take note.
General robustness is currently poorly understood,
and very far from being achieved in practice.
We hope that our work leads to more work
in this area, and to more caution.

\newpage

\bibliography{refs.bib,refs2.bib}

\newpage

\appendix

\newpage

\section{Model details}
\label{app:models}

\subsection{BigGAN}

We use the BigGAN-deep generator architecture
at the $512 \times 512$ resolution,
which can be found in Table 9 of Appendix~B in
the paper introducing BigGAN \cite{Brock:LSG:2019}.
Conveniently, this table clearly indicates the
locations at which we perturb the activations;
every horizontal line of the table is a point
at which our method can perturb the activation values.
Please refer to Appendix~B of
the BigGAN paper for detailed descriptions,
in particular of the ResBlocks which
comprise the majority of the network.
Note that we in essence perform perturbations
after each ResBlock; if desired,
perturbations could also be performed
within each block.
We take the pre-trained generator published
by DeepMind \cite{DeepMind:BD5:2019}.

\subsection{Standard classifiers}

We use two classifiers trained as usual to maximise
accuracy on the training distribution.
The first 
is from the state-of-the-art EfficientNet 
family \cite{DBLP:conf/icml/TanL19}, 
enhanced using noisy student training 
\cite{DBLP:journals/corr/abs-1911-04252}.
We use the best readily available model and pre-trained weights 
for Pytorch, EfficientNet-B4 (Noisy Student)
from \citet{melas-kyriazi_lukemelasefficientnet-pytorch_2020}.
The second 
is PyTorch's pre-trained ResNet50 
\cite{DBLP:conf/cvpr/HeZRS16},
made available through the \texttt{torchvision} package of PyTorch.
These classifiers' ImageNet
accuracies are reported in Table~\ref{tab:classacc}.

\begin{table}[h]
	\centering
	\caption{Classifiers' accuracy on ImageNet, in \%}
	\label{tab:classacc}
	\begin{tabular}{lll}
		\toprule
	Classifier         & Top-1 & Top-5 \\ \midrule
	ResNet50           & 76.15 & 92.87 \\
	EfficientNet-B4 NS & 85.16 & 97.47 \\ \bottomrule
	\end{tabular}
\end{table}

\subsection{Robust classifiers}

We use two pre-trained ResNet50 classifiers adversarially trained 
against bounded pixel perturbations. 
The first, `ResNet50 Robust (Engstrom)', from \citet{robustness}, was
trained using $l_2$-norm projected gradient descent attack with
$\eps = 0.3$. The second, `ResNet50 Robust (``Fast'')',
from \textit{Fast Is Better Than Free: 
	Revisiting Adversarial Training}
 \cite{DBLP:conf/iclr/WongRK20},
was trained with the fast gradient sign method attack
for robustness against
$l_{\infty}$ with $\eps = 4/255$. The classifiers' ImageNet
accuracies and robustness to relevant attacks are shown in 
Table~\ref{tab:robustclassacc}.

\begin{table*}[h]
	\centering
	\caption{Classifiers' accuracy on ImageNet, and robustness to attacks, in \%}
	\label{tab:robustclassacc}
	\begin{tabular}{b{0.3\textwidth}b{0.12\textwidth}b{0.18\textwidth}b{0.23\textwidth}}
		\toprule
	Classifier & Top-1 \newline (no attack) & Top-1 \newline ($l_2$ attack $\eps=0.3$) & Top-1 \newline ($l_\infty$ attack $\eps=4/255$) \\ \midrule
	ResNet50 Robust (Engstrom) & 57.90 & 35.16 & \\
	ResNet50 Robust (``Fast'') & 55.45 &  & 30.28 \\ \bottomrule
	\end{tabular}
\end{table*}

\section{Experimental setup}
\label{app:details}

\subsection{Technical details}
For our experiments, we used used the neural
networks described in Appendix~\ref{app:models},
and searched for context-sensitive perturbations using the procedure
described in Sections~\ref{sec:methods} and \ref{sec:results}.
However, those sections did not describe the optimisation
procedure used.
We used the Adam optimiser \cite{Kingma:AAM:2015} with
a learning rate of 0.03 and the default $\beta$ hyperparameters
of 0.9 and 0.999.
After each optimisation step, we constrained the magnitude of
the perturbation by finding the $L_2$ norm of the perturbation
`vector' obtained by concatenating the scalars used to perturb
each individual activation value, then rescaling it to have
a norm no greater than our constraint.
This constraint was initially set to be magnitude 1,
and was slightly relaxed after each optimisation step
by multiplication by 1.03 and addition of 0.1.
These values were empirically found---using small amount
of manual experimentation---to be a reasonable
tradeoff between starting small and increasing slowly enough
to find decently small perturbations, while also using
a reasonable amount of compute.
Typically, finding a perturbation under this
regime takes $O(100)$ steps, which took
$O(1 \text{ minute})$ using the single
NVIDIA Tesla V100 GPU we used.

\subsection{Data collection}
\label{app:prod_graphs}

To run our ImageNet experiments, 
we began by randomly sampling $(y, z, t)$ tuples,
where $y$ is the desired true image label, $z$ is the latent
input to the generator, and $t$ is the target label for the
perturbed misclassification.
For each classifier, and each kind of perturbation, 
we calculated perturbations based on each of these tuples.
That is, with generator $g$, we perturb the 
original image $g(z;y)$ to be classified as $t$.
These same tuples were used 
for all of the experiments, although the number of such tuples 
in each experiment varies slightly, and can be seen in 
Table~\ref{tab:exp_ns}.

The images then go through two filters, and they 
are not used if they fail to meet the requirements.
Firstly, the classifier's prediction must be $y$, 
the label that was used to produce the image. If not, then 
the classifier is already incorrect, and so it is not worth 
perturbing the image to cause a misclassification. Secondly,
human labellers vote whether the original image was actually 
of class $y$ (this because generators are not perfect, and so
may fail at making an image of the intended class). A majority 
vote is used, and images which are voted to not be of the intended class 
are discarded. The majority vote is similar to that used in the original 
ImageNet labelling process.

Finally, human labellers are asked to determine whether the perturbed 
image is of the same class as the original image, i.e. if the 
perturbation successfully preserved the true class, while changing the 
classifier's predicted class. Again, this is done by a majority vote.

The solid lines in Figure~\ref{fig:graphs} show the final results. Note 
that they do not always reach 1. This is because we consider that if a perturbation 
changed the true class of an image, no suitable perturbation was found.

We separate the images into several
experiments of 30 images and report the mean and 
standard deviation over these experiments.

\begin{table}[h]
	\centering
	\caption{For each classifier, and for each section in which 
	we made perturbations, the number of images included in our final 
	results (Figure~\ref{fig:graphs}). This number does not include
	examples for which the unperturbed image was judged by labellers to be
	of the wrong class; this number is included in parentheses.}
	\label{tab:exp_ns}
	\begin{tabular}{cccc}
		\toprule
		\multicolumn{4}{l}{\textbf{ResNet 50}}                                                    \\
		\midrule
		First 6              & Middle 6             & Last 6               & All 18               \\
		148 (61)             & 147 (62)             & 148 (61)             & 148 (61)             \\
		\bottomrule
		\multicolumn{1}{l}{} & \multicolumn{1}{l}{} & \multicolumn{1}{l}{} & \multicolumn{1}{l}{} \\
		\toprule
		\multicolumn{4}{l}{\textbf{EfficientNet-B4NS}}                                            \\
		\midrule
		First 6              & Middle 6             & Last 6               & All 18               \\
		158 (54)             & 157 (55)             & 161 (52)             & 148 (49)             \\
		\bottomrule
		\multicolumn{1}{l}{} & \multicolumn{1}{l}{} & \multicolumn{1}{l}{} & \multicolumn{1}{l}{} \\
		\toprule
		\multicolumn{4}{l}{\textbf{ResNet50 Robust ("Engstrom")}}                                 \\
		\midrule
		First 6              & Middle 6             & Last 6               & All 18               \\
		136 (51)             & 137 (50)             & 138 (50)             & 133 (55)             \\
		\bottomrule
		\multicolumn{1}{l}{} & \multicolumn{1}{l}{} & \multicolumn{1}{l}{} & \multicolumn{1}{l}{} \\
		\toprule
		\multicolumn{4}{l}{\textbf{ResNet50 Robust ("Fast")}}                                     \\
		\midrule
		First 6              & Middle 6             & Last 6               & All 18               \\
		95 (69)              & 123 (43)             & 123 (43)             & 119 (39)             \\
		\bottomrule
	\end{tabular}
\end{table}

\subsection{Labelling interface}

\begin{figure*}[h]
	\centering
	\frame{\includegraphics[width=\linewidth]{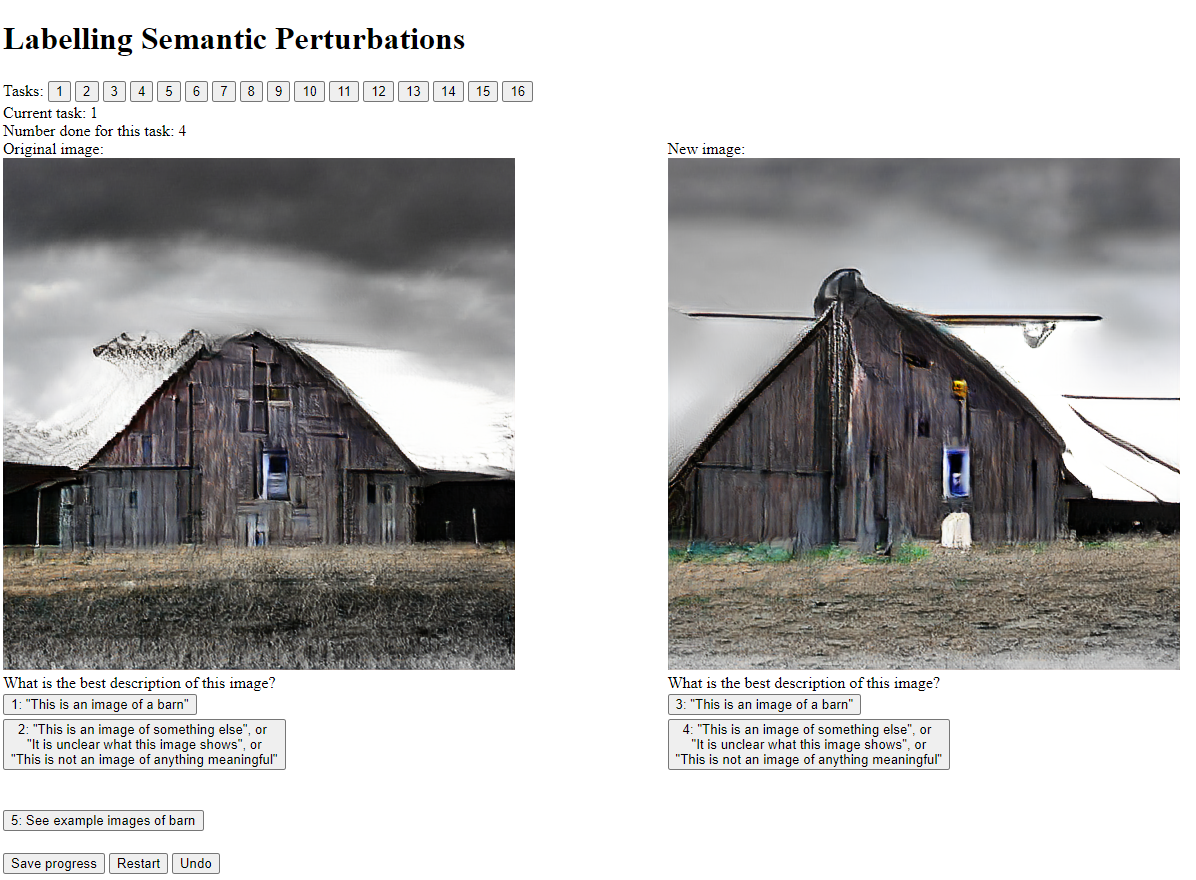}}
	\caption{Screenshot of labelling interface. The perturbed
	image and buttons, on the right-hand side,
	are visible only when the unperturbed image (on the left)
	has been selected as matching the desired label.
	The buttons are numbered to provide keyboard shortcuts.
	The button at the bottom opens a web image search, in
	case the user is unfamiliar with the class label.}
	\label{fig:labelinterface}
\end{figure*}

For both the stages at which human labelling is required,
we use the interface shown in Figure~\ref{fig:labelinterface}.
The user first labels whether the original image is the of 
the intended label, and then the perturbed image. 
We ask the user which of the following four options
is the best description of the image:
\begin{enumerate}
	\item ``This is an image of label $y$''
	\item ``This is an image of something else''
	\item ``It is unclear what this image shows''
	\item ``This is not an image of anything meaningful''
\end{enumerate}

\section{Experiments with no human labelling}
\label{app:no-human}
\begin{figure*}
	\centering
	\begin{subfigure}[t]{.48\linewidth}
		\centering
		\includegraphics[width=\linewidth]{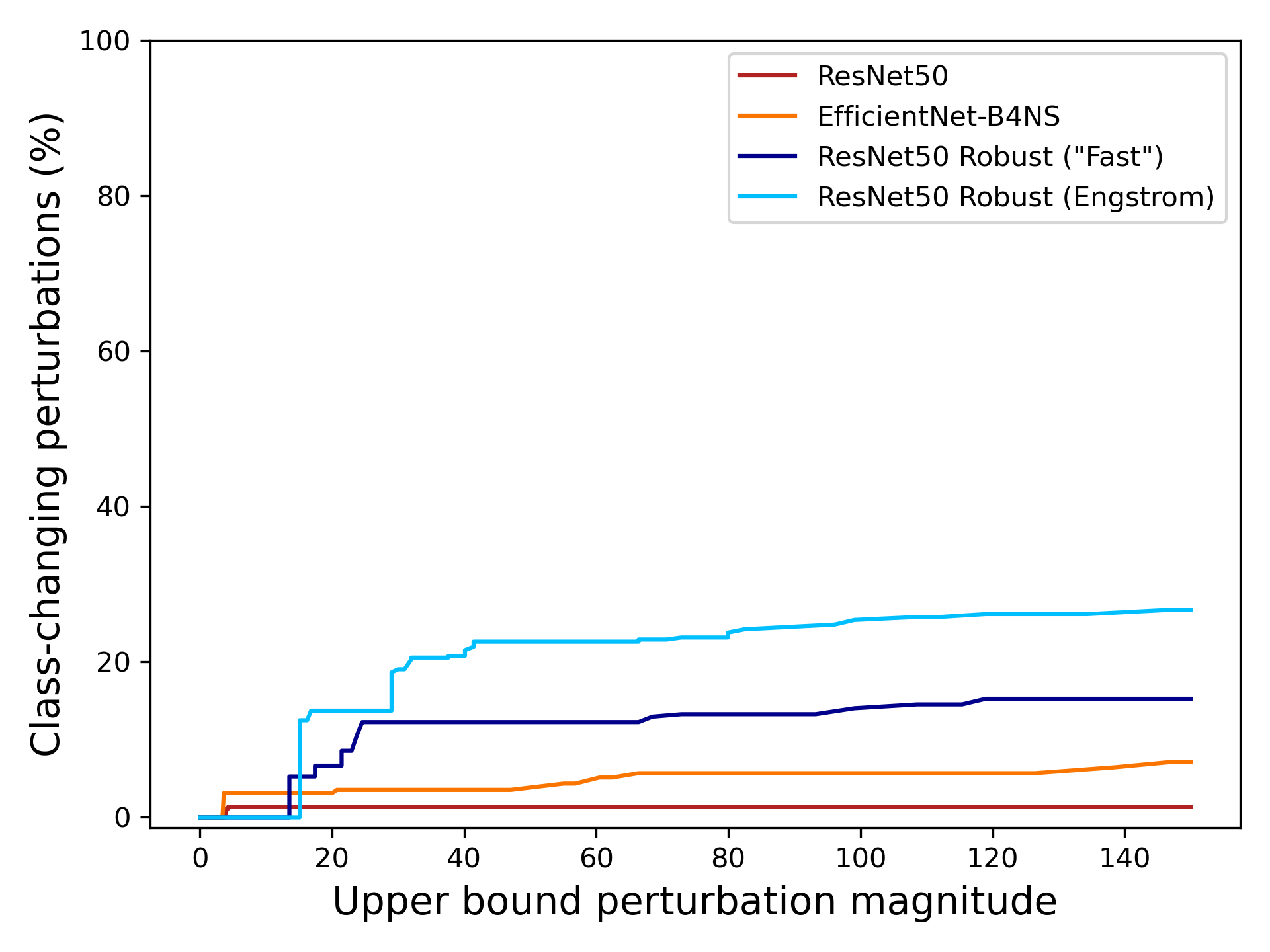}
		\caption{Activation values perturbed at all BigGAN layers.}
		\label{fig:All_layers}
	\end{subfigure}\hfill%
	\begin{subfigure}[t]{.48\linewidth}
		\centering
		\includegraphics[width=\linewidth]{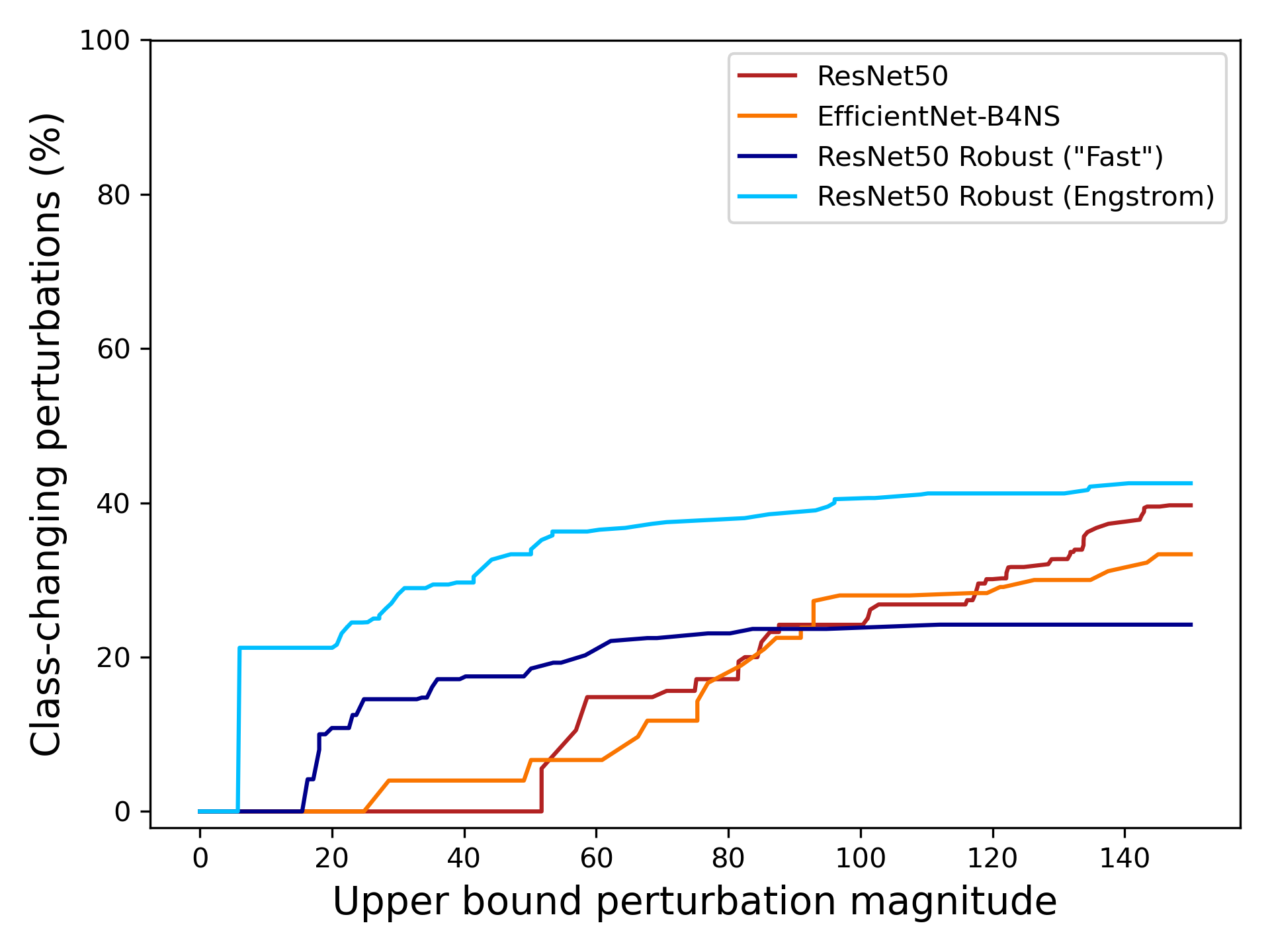}
		\caption{Activation values perturbed in the first six layers only.}
		\label{fig:1/3}
	\end{subfigure}
	
	\vspace{3mm}
	
	\begin{subfigure}[t]{.48\linewidth}
		\centering
		\includegraphics[width=\linewidth]{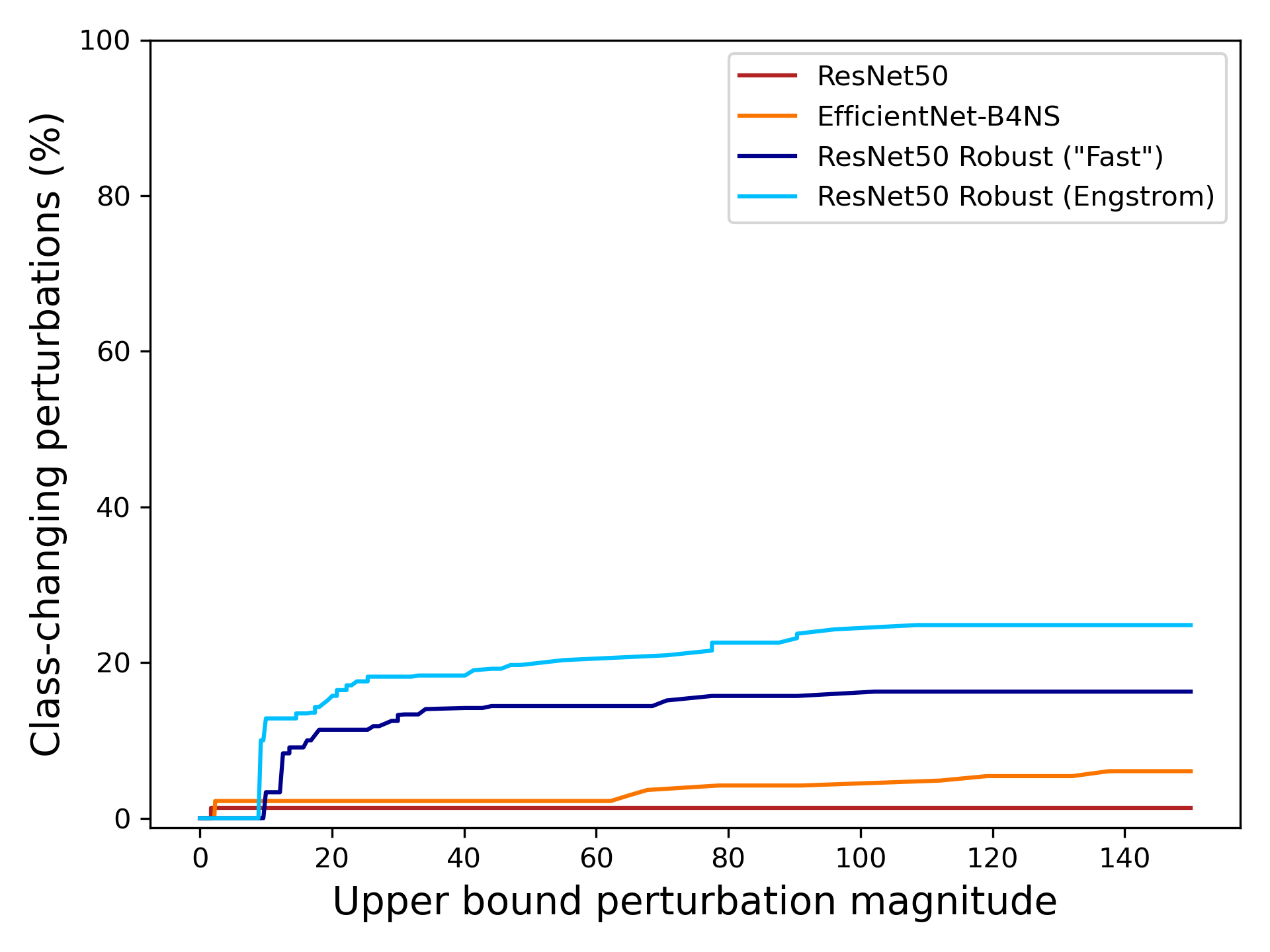}
		\caption{Activations perturbed in the middle six layers only.}
		\label{fig:2/3}
	\end{subfigure}\hfill%
	\begin{subfigure}[t]{.48\linewidth}
		\centering
		\includegraphics[width=\linewidth]{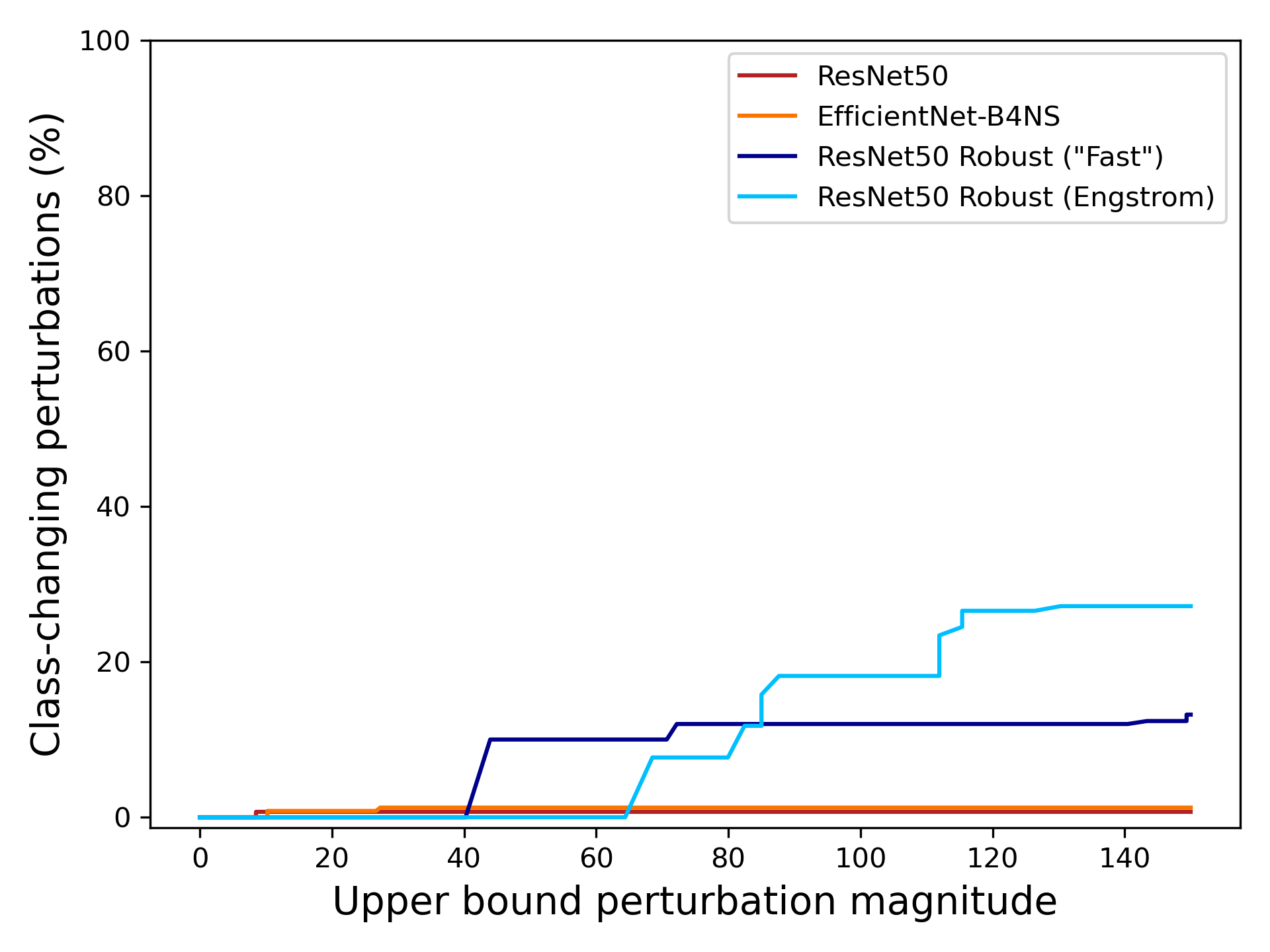}
		\caption{Activation values perturbed in the last six layers only.}
		\label{fig:3/3}
	\end{subfigure}\hfill
	\caption{Number of class-changing perturbations for 
	different upper bounds on perturbation magnitude,
	according to classifier, and where the perturbation is 
	being made.}
	\label{fig:nohumans}
\end{figure*}

Our perturbation method, like any making large 
visual changes to images, has the potential to change 
the true class of the image. We tackled this problem 
by using human labellers to identify the cases in which 
this happens. However, it is also possible to 
use the standard approach of constraining 
the perturbation magnitude, to limit the visual change 
in the final image. By setting an upper bound on the perturbation 
magnitude, we are more constrained 
in the changes we can make to images,
but we also make fewer class-changing perturbations. 
This allows us to avoid human labelling, by accepting a
small amount of error in the labels. The only way to completely
avoid this error would be to make changes that \emph{cannot} change
the true class (very limiting), or have humans label the images.
Figure~\ref{fig:nohumans} demonstrates this 
tradeoff for our ImageNet experiments.

\section{ImageNet: further examples}

\label{app:imagenet-examples}

See Figures~\ref{fig:standard_version_5} and 
\ref{fig:robust_version_5}
below, which give some
examples of context-sensitive feature perturbations.

Please also refer to our further appendices,
which can be found at \url{https://doi.org/10.5281/zenodo.4121010
}.
These include many more ImageNet examples,
as well as many
animations showing the effect of gradually
introducing the perturbations to the latent
activations of the generator.
These give a much clearer intuition for
the nature of the changes being made to the images;
comparing static images alone can be difficult
to interpret.

\setlength{\mypicwidth}{.175\textwidth}

\begin{figure*}[]
	\centering
	\begin{tabular}[t]{ccccc}
			\multicolumn{2}{l}{\Large{ResNet50}} &
			\multicolumn{3}{r}{\Large{ `palace' $\rightarrow$ `throne'}} \\
			&& ($\times$10 for visibility) & ($\times$25 for visibility) & ($\times$5 for visibility) \\
			\frame{{\includegraphics[width=\mypicwidth]{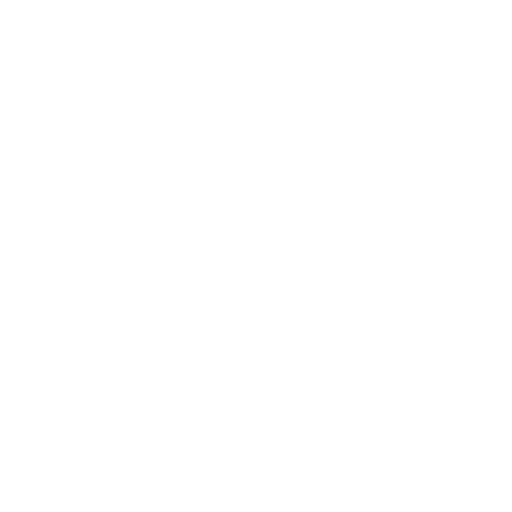}}} &
			\frame{\includegraphics[width=\mypicwidth]{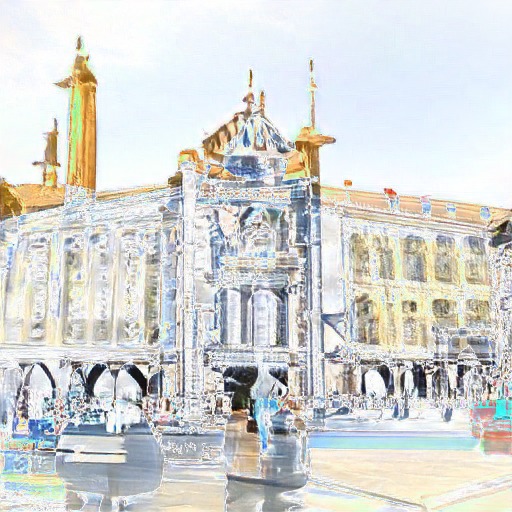}} &
			\frame{\includegraphics[width=\mypicwidth]{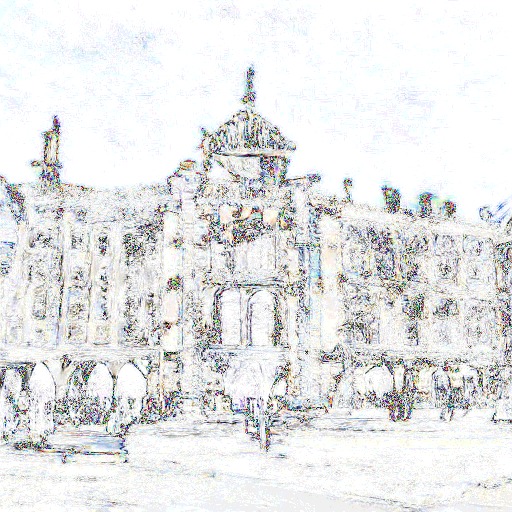}} &
			\frame{\includegraphics[width=\mypicwidth]{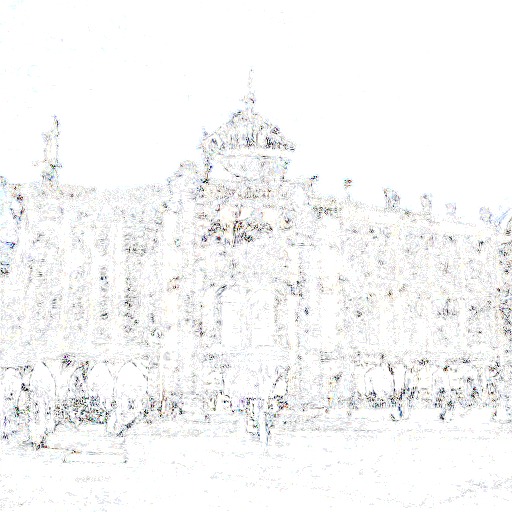}} &
			\frame{\includegraphics[width=\mypicwidth]{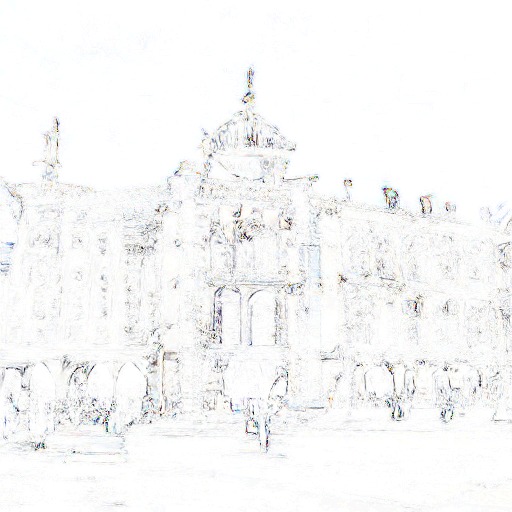}} \\
			\frame{\includegraphics[width=\mypicwidth]{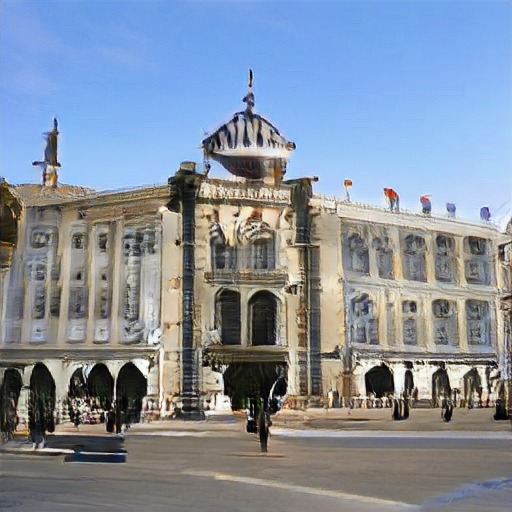}} &
			\frame{\includegraphics[width=\mypicwidth]{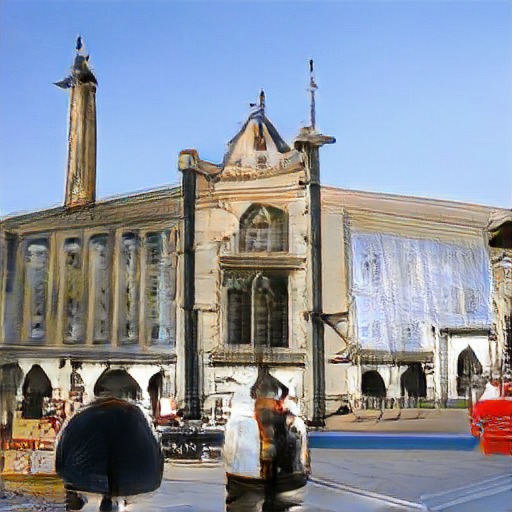}} &
			\frame{\includegraphics[width=\mypicwidth]{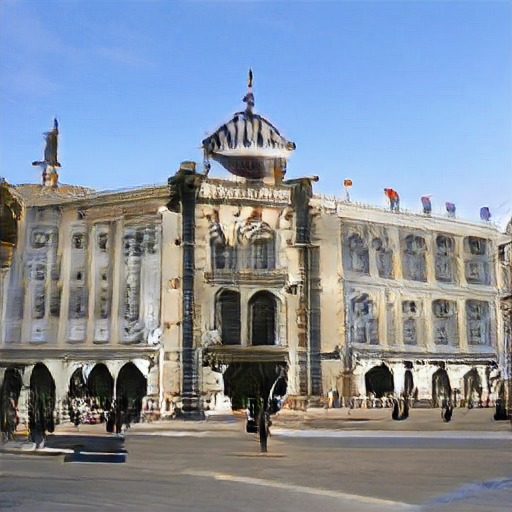}} &
			\frame{\includegraphics[width=\mypicwidth]{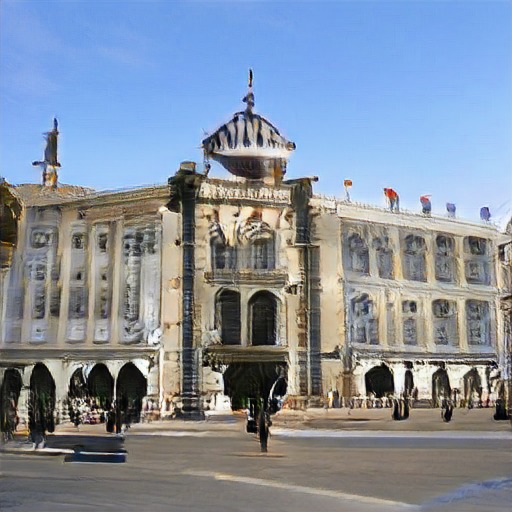}} &
			\frame{\includegraphics[width=\mypicwidth]{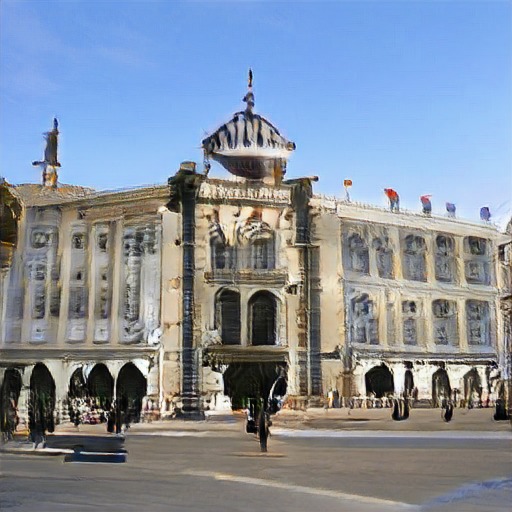}} \\
			original & first 6 layers & middle 6 layers & last 6 layers & all 18 layers \vspace{1cm} \\
			\multicolumn{2}{l}{\Large{EfficientNet-B4NS}} &
			\multicolumn{3}{r}{\Large{ `palace' $\rightarrow$ `throne'}} \\
			&& ($\times$10 for visibility) & ($\times$25 for visibility) & ($\times$5 for visibility) \\
			\frame{{\includegraphics[width=\mypicwidth]{{imagenet_appendix/whitebox.jpg}}}} &
			\frame{\includegraphics[width=\mypicwidth]{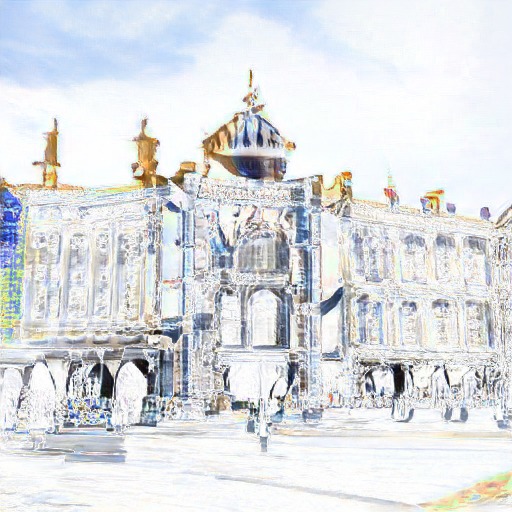}} &
			\frame{\includegraphics[width=\mypicwidth]{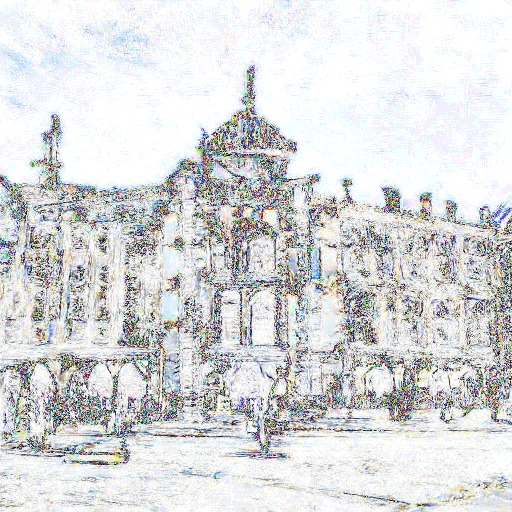}} &
			\frame{\includegraphics[width=\mypicwidth]{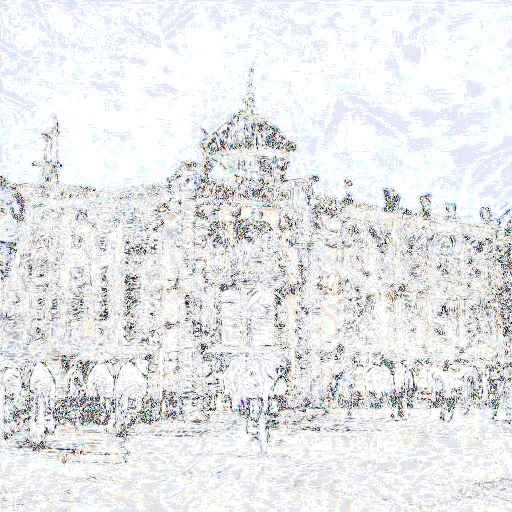}} &
			\frame{\includegraphics[width=\mypicwidth]{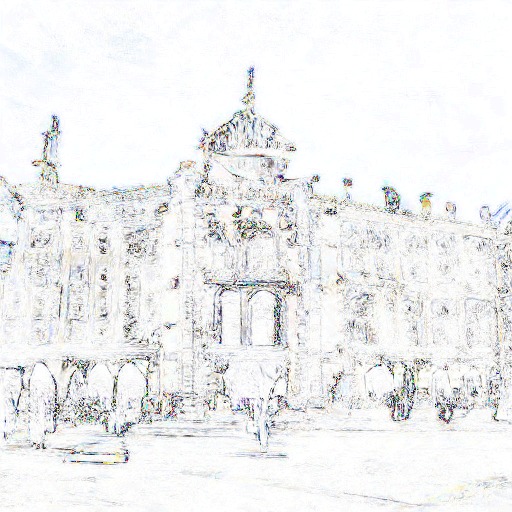}} \\
			\frame{\includegraphics[width=\mypicwidth]{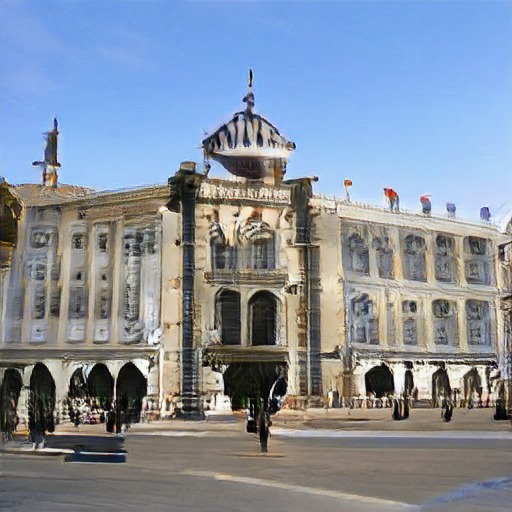}} &
			\frame{\includegraphics[width=\mypicwidth]{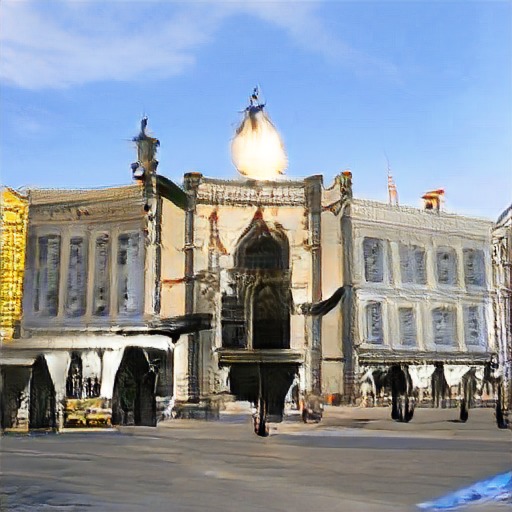}} &
			\frame{\includegraphics[width=\mypicwidth]{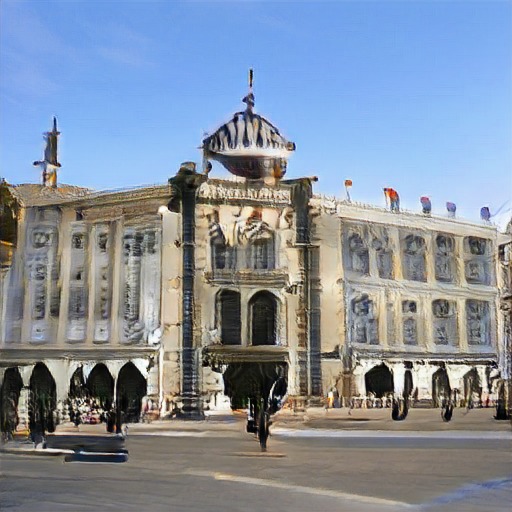}} &
			\frame{\includegraphics[width=\mypicwidth]{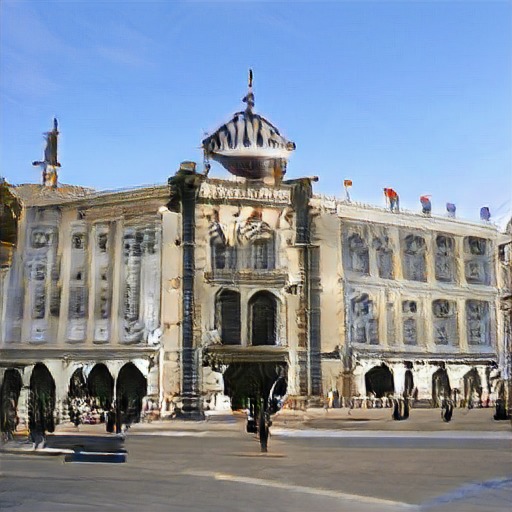}} &
			\frame{\includegraphics[width=\mypicwidth]{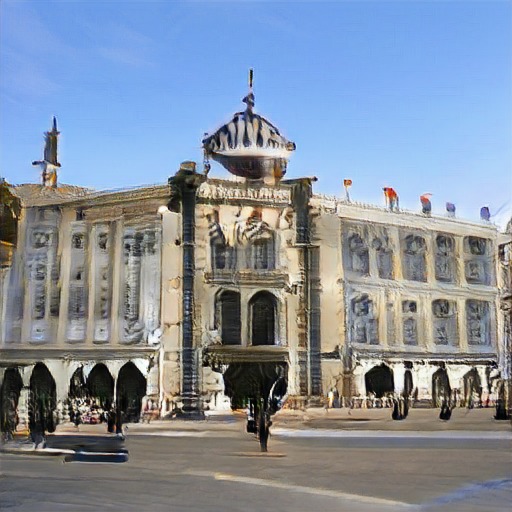}} \\
			original & first 6 layers & middle 6 layers & last 6 layers & all 18 layers \vspace{1cm} \\
		\end{tabular}
		\caption{Examples of feature perturbations for the two standard classifiers.
		For each, the bottom row show the perturbed images 
		for perturbations at different parts of the generator. The top row
		shows the pixel-wise difference between the original image and 
		the perturbed image. Some of these have been scaled to be made more visible.
		The name of the classifier is shown in the top left, and in the top right,
		the original and target label.}
		\label{fig:standard_version_5}
	\end{figure*}
	
	\begin{figure*}[]
		\centering
		\begin{tabular}[t]{ccccc}
			\multicolumn{2}{l}{\Large{ResNet50 Robust (``Engstrom")}} &
			\multicolumn{3}{r}{\Large{ `palace' $\rightarrow$ `throne'}} \\
			&&&& \\
			\frame{{\includegraphics[width=\mypicwidth]{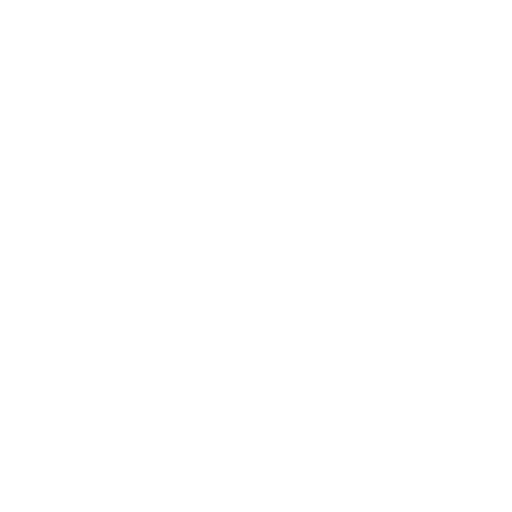}}} &
			\frame{\includegraphics[width=\mypicwidth]{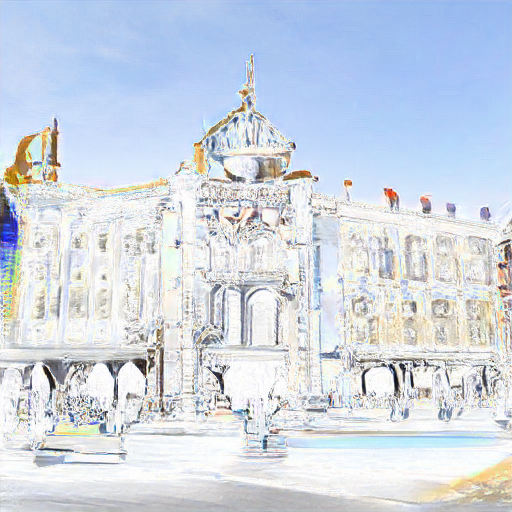}} &
			\frame{\includegraphics[width=\mypicwidth]{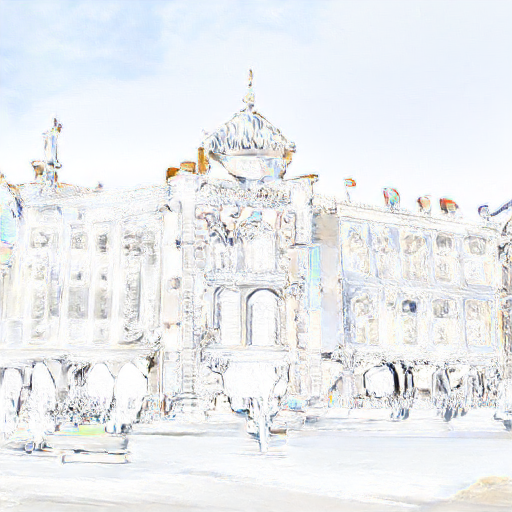}} &
			\frame{\includegraphics[width=\mypicwidth]{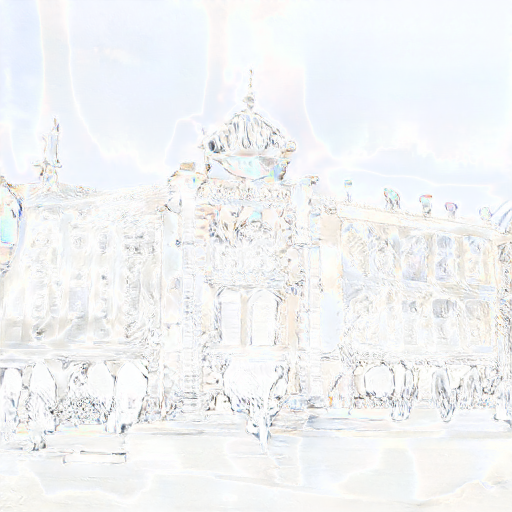}} &
			\frame{\includegraphics[width=\mypicwidth]{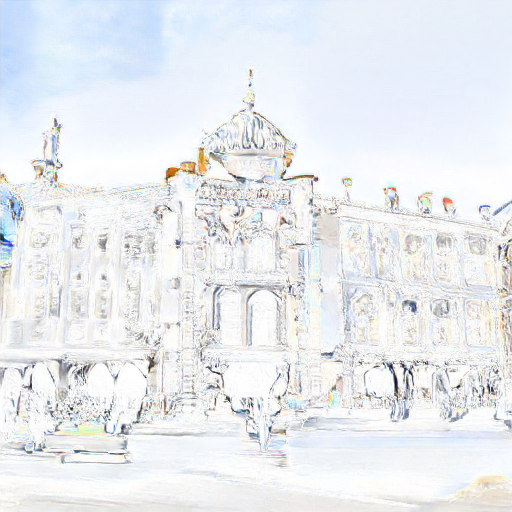}} \\
			\frame{\includegraphics[width=\mypicwidth]{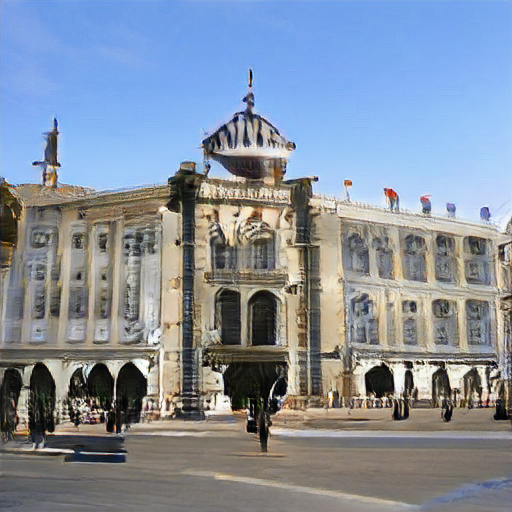}} &
			\frame{\includegraphics[width=\mypicwidth]{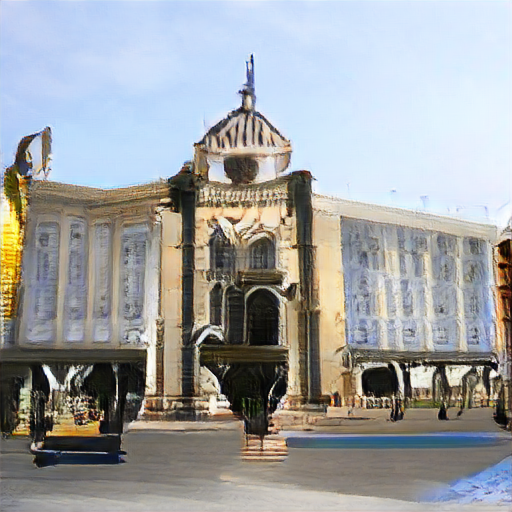}} &
			\frame{\includegraphics[width=\mypicwidth]{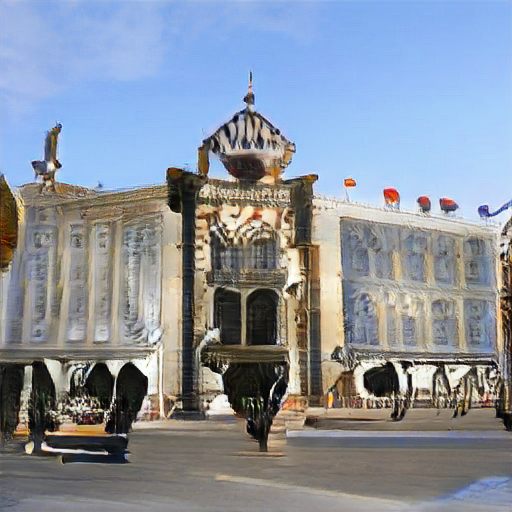}} &
			\frame{\includegraphics[width=\mypicwidth]{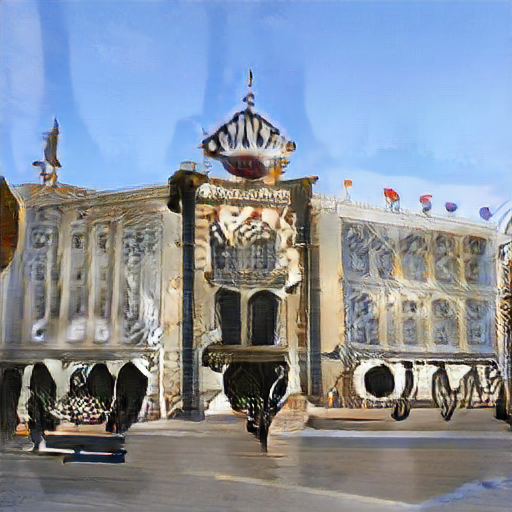}} &
			\frame{\includegraphics[width=\mypicwidth]{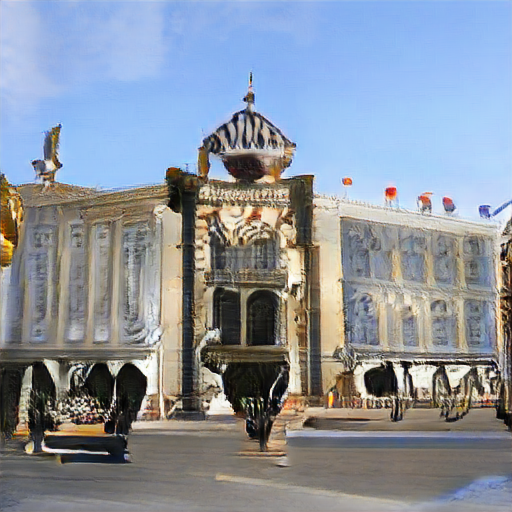}} \\
			original & first 6 layers & middle 6 layers & last 6 layers & all 18 layers \vspace{1cm} \\
			\multicolumn{2}{l}{\Large{ResNet50 Robust (``Fast")}} &
			\multicolumn{3}{r}{\Large{ `palace' $\rightarrow$ `throne'}} \\
			&&&& \\
			\frame{{\includegraphics[width=\mypicwidth]{{imagenet_appendix/whitebox.png}}}} &
			\frame{\includegraphics[width=\mypicwidth]{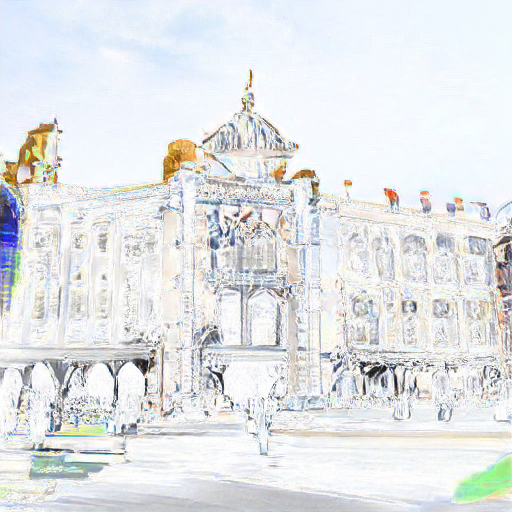}} &
			\frame{\includegraphics[width=\mypicwidth]{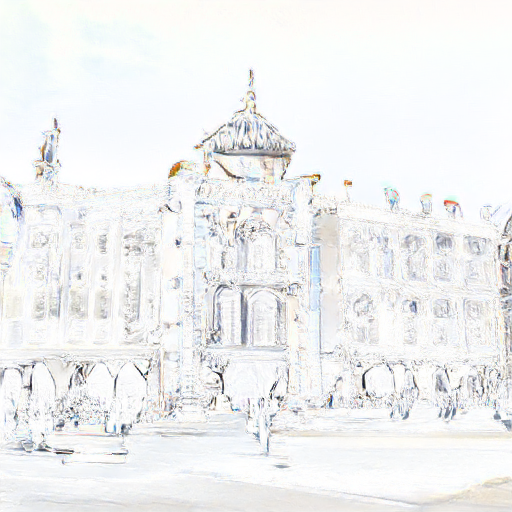}} &
			\frame{\includegraphics[width=\mypicwidth]{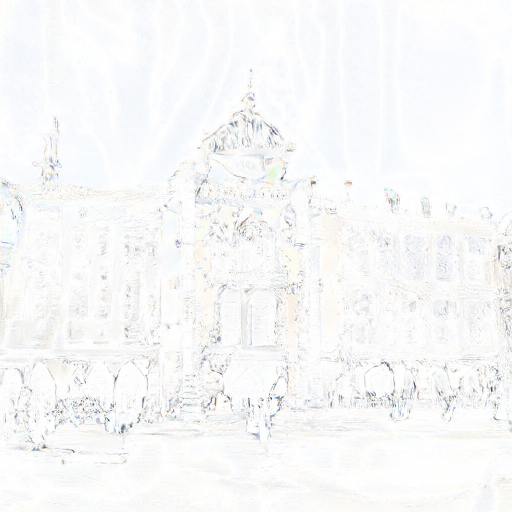}} &
			\frame{\includegraphics[width=\mypicwidth]{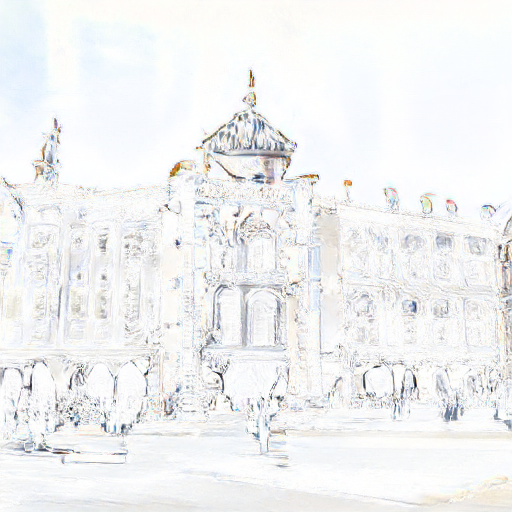}} \\
			\frame{\includegraphics[width=\mypicwidth]{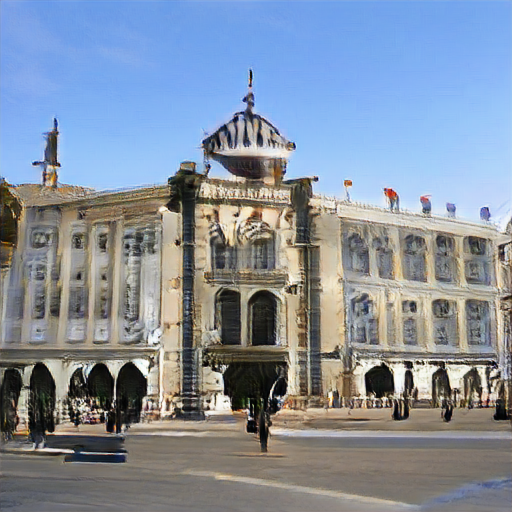}} &
			\frame{\includegraphics[width=\mypicwidth]{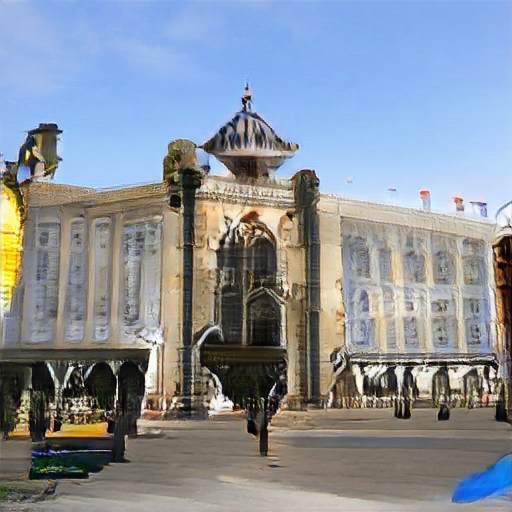}} &
			\frame{\includegraphics[width=\mypicwidth]{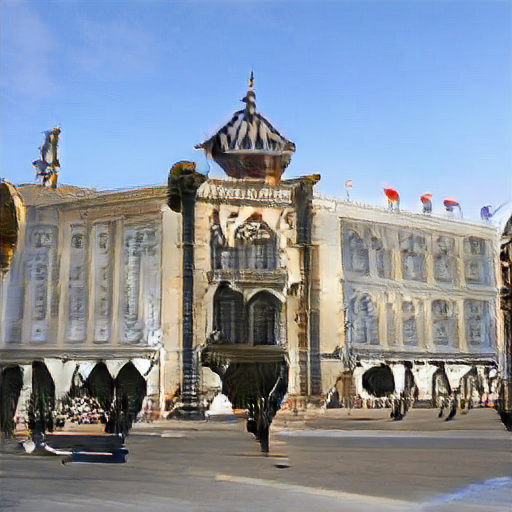}} &
			\frame{\includegraphics[width=\mypicwidth]{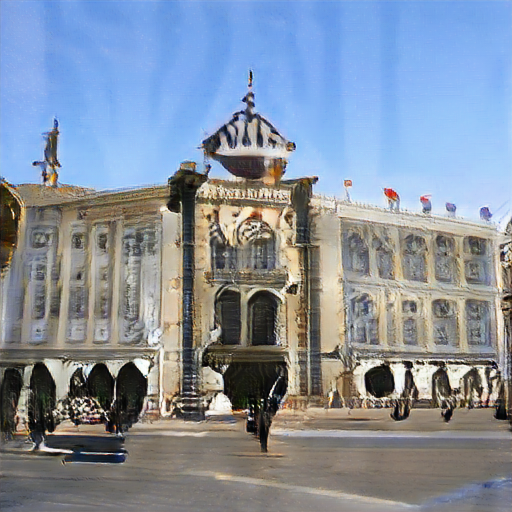}} &
			\frame{\includegraphics[width=\mypicwidth]{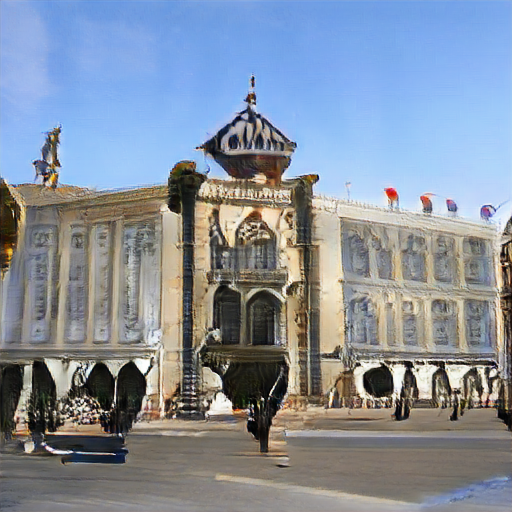}} \\
			original & first 6 layers & middle 6 layers & last 6 layers & all 18 layers \vspace{1cm} \\
		\end{tabular}
		\caption{Examples of feature perturbations for the two pixel-robust classifiers.
		For each, the bottom row show the perturbed images 
		for perturbations at different parts of the generator. The top row
		shows the pixel-wise difference between the original image and 
		the perturbed image.
		The name of the classifier is shown in the top left, and in the top right,
		the original and target label.}
		\label{fig:robust_version_5}
	\end{figure*}

\clearpage

\section{CelebA-HQ}
\label{app:celeba}
\subsection{Model details}
\paragraph{Progressive GAN}
We use the pretrained CelebA-HQ $512 \times 512$ Progressive GAN from
\url{https://pytorch.org/hub/facebookresearch_pytorch-gan-zoo_pgan/}.
We simply perturb the activations after each \backtick scaleLayer'
in this implementation.
Note that unlike the other generative models we use,
this is not a conditional model. That is, its only input
is the random seed: you cannot specify that it generates an
image with certain characteristics.

Table \ref{tab:celeba-gen} details the layers of the Progressive GAN,
and indicates which activations are perturbed.

\begin{table*}
	\caption{CelebA-HQ convolutional generator architecture.
		Each row represents
		a layer. Each horizontal rule marks an
		activation tensor at which perturbations are performed.}
	\label{tab:celeba-gen}
	\centering
	\begin{tabular}{rll}
		\multirow{2}{*}{} & & \\ \cline{2-3}	&&\\
		& Fully-Connected &  (8192 units) \\
		& LeakyReLU &  (Slope $-0.2$) \\
		& Reshape &  (To batch of $512 \times 4 \times 4$ tensors) \\
		& 2D Convolution & ($3 \times 3$ kernel, stride 1, padding size 1, 512 feature maps) \\
		& LeakyReLU &  (Slope $-0.2$) \\
		
		\multirow{2}{*}{} & & \\ \cline{2-3}	&&\\
		
		& Upscale & (To $8 \times 8$) \\
		& 2D Convolution & ($3 \times 3$ kernel, stride 1, padding size 1, 512 feature maps) \\
		& LeakyReLU &  (Slope $-0.2$) \\
		& 2D Convolution & ($3 \times 3$ kernel, stride 1, padding size 1, 512 feature maps) \\
		& LeakyReLU &  (Slope $-0.2$) \\
		
		\multirow{2}{*}{} & & \\ \cline{2-3}	&&\\
		
		& Upscale & (To $16 \times 16$) \\
		& 2D Convolution & ($3 \times 3$ kernel, stride 1, padding size 1, 512 feature maps) \\
		& LeakyReLU &  (Slope $-0.2$) \\
		& 2D Convolution & ($3 \times 3$ kernel, stride 1, padding size 1, 512 feature maps) \\
		& LeakyReLU &  (Slope $-0.2$) \\
		
		\multirow{2}{*}{} & & \\ \cline{2-3}	&&\\
		
		& Upscale & (To $32 \times 32$) \\
		& 2D Convolution & ($3 \times 3$ kernel, stride 1, padding size 1, 512 feature maps) \\
		& LeakyReLU &  (Slope $-0.2$) \\
		& 2D Convolution & ($3 \times 3$ kernel, stride 1, padding size 1, 512 feature maps) \\
		& LeakyReLU &  (Slope $-0.2$) \\
		
		\multirow{2}{*}{} & & \\ \cline{2-3}	&&\\
		
		& Upscale & (To $64 \times 64$) \\
		& 2D Convolution & ($3 \times 3$ kernel, stride 1, padding size 1, 256 feature maps) \\
		& LeakyReLU &  (Slope $-0.2$) \\
		& 2D Convolution & ($3 \times 3$ kernel, stride 1, padding size 1, 256 feature maps) \\
		& LeakyReLU &  (Slope $-0.2$) \\
		
		\multirow{2}{*}{} & & \\ \cline{2-3}	&&\\
		
		& Upscale & (To $128 \times 128$) \\
		& 2D Convolution & ($64 \times 64$ kernel, stride 1, padding size 1, 128 feature maps) \\
		& LeakyReLU &  (Slope $-0.2$) \\
		& 2D Convolution & ($3 \times 3$ kernel, stride 1, padding size 1, 128 feature maps) \\
		& LeakyReLU &  (Slope $-0.2$) \\
		
		\multirow{2}{*}{} & & \\ \cline{2-3}	&&\\
		
		& Upscale & (To $256 \times 256$) \\
		& 2D Convolution & ($3 \times 3$ kernel, stride 1, padding size 1, 64 feature maps) \\
		& LeakyReLU &  (Slope $-0.2$) \\
		& 2D Convolution & ($3 \times 3$ kernel, stride 1, padding size 1, 64 feature maps) \\
		& LeakyReLU &  (Slope $-0.2$) \\
		
		\multirow{2}{*}{} & & \\ \cline{2-3}	&&\\
		
		& Upscale & (To $512 \times 512$) \\
		& 2D Convolution & ($3 \times 3$ kernel, stride 1, padding size 1, 32 feature maps) \\
		& LeakyReLU &  (Slope $-0.2$) \\
		& 2D Convolution & ($3 \times 3$ kernel, stride 1, padding size 1, 32 feature maps) \\
		& LeakyReLU &  (Slope $-0.2$) \\
		
		\multirow{2}{*}{} & & \\ \cline{2-3}	&&\\
		
		& 2D Convolution & ($1 \times 1$ kernel, stride 1, 3 feature maps) \\
		\multirow{2}{*}{} & & \\ \cline{2-3}	&&\\
	\end{tabular}
\end{table*}

\paragraph{Classifier}
CelebA is used primarily as benchmark for generative modelling,
not discriminative classification. We could not find any
pre-trained classifiers for the 40 binary attributes that the dataset
is labelled with. In the absence of any suitable checkpoints,
we simply used existing code to train the classifier we needed:
\url{https://github.com/aayushmnit/Deep_learning_explorations/tree/master/7_Facial_attributes_fastai_opencv}.
The resulting model obtains $>90\%$ accuracy over the forty binary labels,
certainly good enough for our purpose of demonstrating our method.

\subsection{Experimental setup}
CelebA is labelled with 40 binary attributes.
It is very easy to flip the prediction of just one
of these attribute predictions, but is difficult to
flip all forty at once, if only because this is
a forty-objective optimisation problem;
multi-objective optimisation is notoriously challenging.
As a sensible middle ground, we use our method to find
context-sensitive perturbations that flip the sign of
ten of the forty attributes, since
$2^{10} = 1024$, which is roughly the number of ImageNet classes.
In particular, because the generator is not conditional,
we cannot know which attribute predictions are correct.
Our approach is therefore to perturb each image so that
all the following labels are predicted positively:
\backtick Bald',
\backtick Blond hair',
\backtick Eyeglasses',
\backtick Goatee',
\backtick Grey hair',
\backtick Moustache',
\backtick No beard',
\backtick Wearing hat',
\backtick Wearing necklace',
and \backtick Wearing necktie'.

Since the generator has ten layers, we demonstrate the effects
of perturbing the first four layers only, the next three layers,
the final three layers, and all ten layers at once.
The optimisation process required a modest amount of
finetuning (a few hours of ad-hoc manual experimentation);
as noted elsewhere, we perform \emph{no} tuning of the layers
selected to perturb at, or the relative scales of the
perturbations at different neurons.
We use a learning rate of $0.1$.
No epsilon bound is needed, since using this
learning rate, the optimisation
converges suitably without it.
To help with the multi-objective optimisation,
the logits are raised to the power of $\frac{1}{10}$,
making the gradients steeper for the constraints not yet
satisfied, and disincentivising further optimisation
of the objectives already satisfied.

We did not have the resources to have an independent
judge label these results,
but we are satisfied by inspection that the results
are similar to the ImageNet results, in that the
large majority of perturbed images have not changed
their original labels.
Note that this claim is not about the photorealism of the
generated images---which depends mainly on the generative model
used---but on whether the perturbed images are not
generally either
unrecognisable as faces, or perturbed so that the
predicted labels become accurate.

\subsection{Results}

We provide results in Figures
\ref{fig:celeba1}, \ref{fig:celeba2}
and \ref{fig:celeba3}.

Please also refer to 
our further appendices at \url{https://doi.org/10.5281/zenodo.4121010}
to see animations showing the effect of gradually
introducing the perturbations to the latent
activations of the generator.
These give the viewer a much clearer intuition for
the nature of the changes being made to the images;
using static comparisons alone can be difficult
to interpret.

\newlength{\celebawidth}
\setlength{\celebawidth}{.18\textwidth}

\newcommand{\celebaimg}[2]{\begin{subfigure}[b]{\celebawidth}
		\centering
		#2 %
		\frame{\includegraphics[width=0.95\linewidth]{#1.jpg}}
	\end{subfigure}\hfill%
}

\newcommand{\celebarow}[4]{
	\centering
	\celebaimg{new_advex/volcs_depths/whitebox}{}
	\celebaimg{AAAI_examples/celeba/version_#1/semantic_pert_diffs_grid_0}{}
	\celebaimg{AAAI_examples/celeba/version_#2/semantic_pert_diffs_grid_0}{($\times 5$ for visibility) \\}
	\celebaimg{AAAI_examples/celeba/version_#3/semantic_pert_diffs_grid_0}{($\times 10$ for visibility) \\}
	\celebaimg{AAAI_examples/celeba/version_#4/semantic_pert_diffs_grid_0}{}

	\vspace{0.2cm}

	\celebaimg{AAAI_examples/celeba/version_#1/unpert_generated_x_grid_0}{}
	\celebaimg{AAAI_examples/celeba/version_#1/generated_x_grid_0}{}
	\celebaimg{AAAI_examples/celeba/version_#2/generated_x_grid_0}{}
	\celebaimg{AAAI_examples/celeba/version_#3/generated_x_grid_0}{}
	\celebaimg{AAAI_examples/celeba/version_#4/generated_x_grid_0}{}

}

\begin{figure*}
	\centering
	\celebarow{415}{416}{417}{332}

	\vspace{0.4cm}

	\celebarow{418}{419}{420}{333}

	\vspace{0.4cm}

	\celebarow{421}{422}{423}{334}

	\caption{A random selection of
		context-sensitive feature perturbations
		at different granularities, as controlled
		by perturbing activations at the
		generator layers indicated under each image.
		Differences with the unperturbed 
		image are shown above each perturbed image.
		Each perturbed image has the following labels
		predicted positively:
		\backtick Bald',
		\backtick Blond hair',
		\backtick Eyeglasses',
		\backtick Goatee',
		\backtick Grey hair',
		\backtick Moustache',
		\backtick No beard',
		\backtick Wearing hat',
		\backtick Wearing necklace',
		and \backtick Wearing necktie'.
	}
	\label{fig:celeba1}
\end{figure*}

\begin{figure*}
	\centering
	\celebarow{424}{425}{426}{335}
	
	\vspace{0.4cm}
	
	\celebarow{427}{428}{429}{336}
	
	\vspace{0.4cm}
	
	\celebarow{430}{431}{432}{337}
	
	\caption{A random selection of
		context-sensitive feature perturbations
		at different granularities, as controlled
		by perturbing activations at the
		generator layers indicated under each image.
		Differences with the unperturbed 
		image are shown above each perturbed image.
		Each perturbed image has the following labels
		predicted positively:
		\backtick Bald',
		\backtick Blond hair',
		\backtick Eyeglasses',
		\backtick Goatee',
		\backtick Grey hair',
		\backtick Moustache',
		\backtick No beard',
		\backtick Wearing hat',
		\backtick Wearing necklace',
		and \backtick Wearing necktie'.
	}
	\label{fig:celeba2}
\end{figure*}

\begin{figure*}
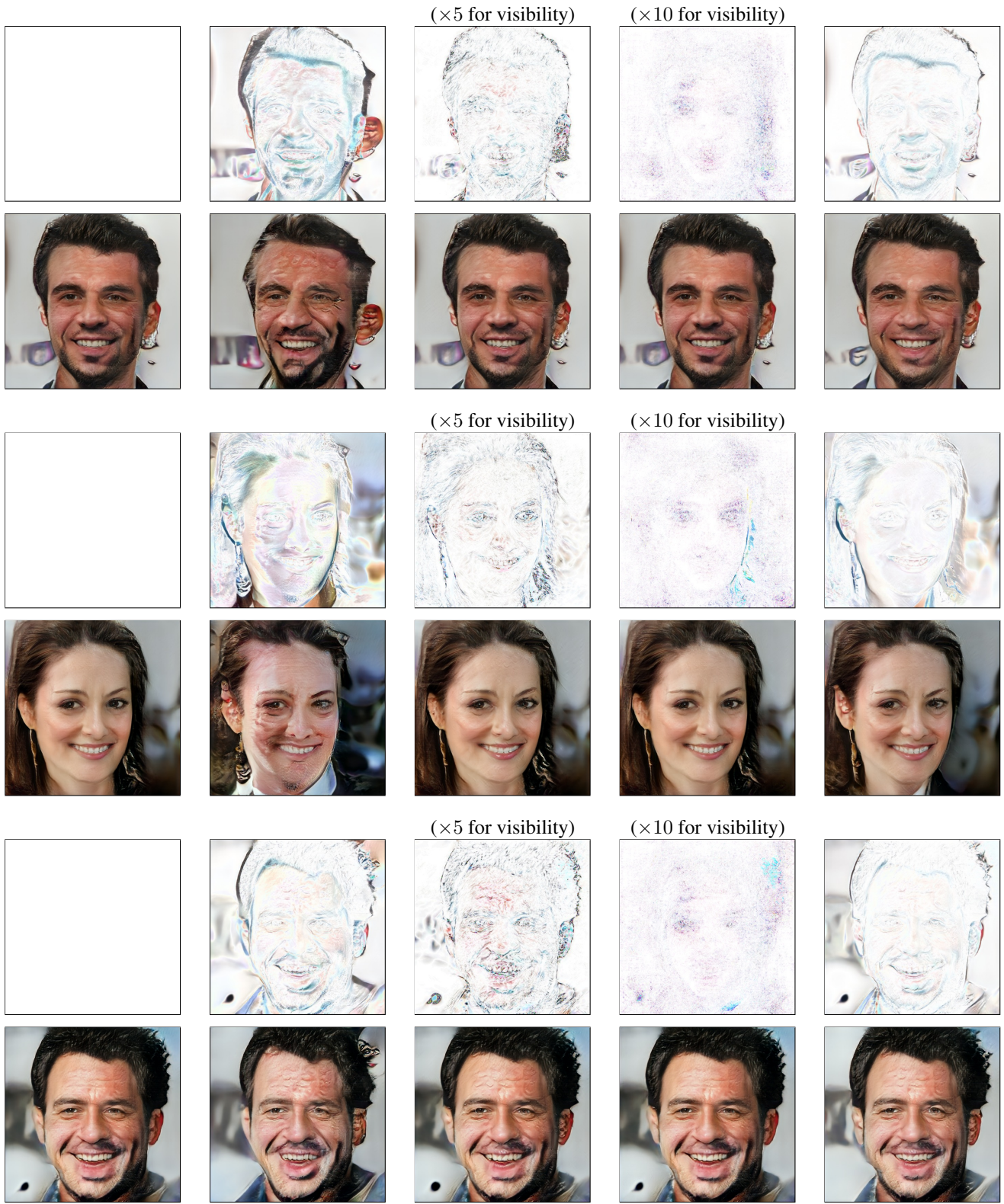

	\centering
	\celebarow{433}{434}{435}{338}
	
	\vspace{0.4cm}
	
	\celebarow{436}{437}{438}{339}
	
	\vspace{0.4cm}
	
	\celebarow{439}{440}{441}{340}
	
	\caption{A random selection of
		context-sensitive feature perturbations
		at different granularities, as controlled
		by perturbing activations at the
		generator layers indicated under each image.
		Differences with the unperturbed 
		image are shown above each perturbed image.
		Each perturbed image has the following labels
		predicted positively:
		\backtick Bald',
		\backtick Blond hair',
		\backtick Eyeglasses',
		\backtick Goatee',
		\backtick Grey hair',
		\backtick Moustache',
		\backtick No beard',
		\backtick Wearing hat',
		\backtick Wearing necklace',
		and \backtick Wearing necktie'.
	}
	\label{fig:celeba3}
\end{figure*}

\clearpage

\section{MNIST}
\label{app:mnist}

\subsection{Model details}

\paragraph{MNIST convolutional GAN}
For MNIST, we tried a range of generators
and found that they all worked roughly as
well as one another.
For the experiments, we use a
simple convolutional generator, inspired
by the Deep Convolutional GAN \cite{Radford:URL:2016}.
Details are shown
in Table~\ref{tab:mnist-gen}.
Inputs to the generator are drawn from a
128-dimensional standard Gaussian.
The sigmoid output transformation ensures
that pixels are in the range
$[0,1]$, as expected by the classifiers.
We perform context-sensitive perturbations
before ReLU layers, rather than
after, to prevent ReLU output
values from being
perturbed to become negative,
which would not have been
encountered during training
and so may not result in
plausible images being generated
since they are out-of-distribution
for the rest of the generator.
Note that perturbing before and
after the
sigmoid transformation has different
effects because
perturbations to values not
close to 0 are diminished in magnitude
if passed through the sigmoid function.

\begin{table}[h]
	\caption{MNIST convolutional generator architecture. Each row represents
		 a layer. Each horizontal rule marks an
		 activation tensor at which perturbations are performed.}
	\label{tab:mnist-gen}
	\centering
\begin{tabular}{lp{35mm}p{35mm}}
	\multirow{2}{*}{} & & \\ \cline{2-3}	&&\\
	 & Fully-Connected &  (64 units) \\
	\multirow{2}{*}{} & & \\ \cline{2-3}	&&\\
	 & ReLU &  \\
	 & Transposed Convolution & ($5 \times 5$ kernel, $2 \times 2$ stride, 32 feature maps) \\
	\multirow{2}{*}{} & & \\ \cline{2-3}	&&\\
	& Batch Normalisation &  \\
	& Leaky ReLU & (Slope $-0.2$)  \\
	& Dropout & ($p = 0.35$)  \\
	& Transposed Convolution & ($5 \times 5$ kernel, $2 \times 2$ stride, 8 feature maps) \\
	\multirow{2}{*}{} & & \\ \cline{2-3}	&&\\
	& Batch Normalisation &  \\
	& Leaky ReLU & (Slope $-0.2$)  \\
	& Dropout & ($p = 0.35$)  \\
	& Transposed Convolution & ($5 \times 5$ kernel, $2 \times 2$ stride, 4 feature maps) \\
	\multirow{2}{*}{} & & \\ \cline{2-3}	&&\\
	& Batch Normalisation &  \\
	& Leaky ReLU & (Slope $-0.2$)  \\
	& Dropout & ($p = 0.35$)  \\
	& Fully-Connected & (784 units) \\
	\multirow{2}{*}{} & & \\ \cline{2-3}	&&\\
	& Sigmoid & ($\tanh$ used during training) \\
	\multirow{2}{*}{} & & \\ \cline{2-3}	&&\\
\end{tabular}
\end{table}

\paragraph{Classifiers}

We use two neural networks that classify MNIST.
One is a simple standard classifier with two convolutional
layers and three fully-connected layers, trained to give
an accuracy over 99\%.
The other is an LeNet5 classifier adversarially 
trained against $l_2$-norm bounded perturbations 
for $\eps=0.3$. This was trained using the 
AdverTorch library \cite{ding2019advertorch}.
It has a standard accuracy of 98\%, reduced to 
95\% by an $l_2$-norm bounded attack 
with $\eps=0.3$.

\subsection{Experimental setup}
We find context-sensitive feature perturbations as described for ImageNet
in Appendix~\ref{app:details}, with a few differences.
First, since the generator is much smaller, we divide it
nearly in half, comparing the effect of perturbing the
activation values at first four layers only
with the effect of perturbing at the last four layers only.
Second, because MNIST (and the generator) is much smaller,
so are the perturbation magnitudes required. So the
learning rate is reduced to 0.004, the
we start with an initial perturbation magnitude constraint of
0.1, and this is gradually relaxed after each optimisation
step by increasing this upper bound by 0.001.
The procedure for collecting human judgements is as described
in Appendix~\ref{app:details}.

\subsection{Results and discussion}
\begin{figure}
	\centering
	\begin{subfigure}[t]{.49\textwidth}
		\centering
		\includegraphics[width=\linewidth]{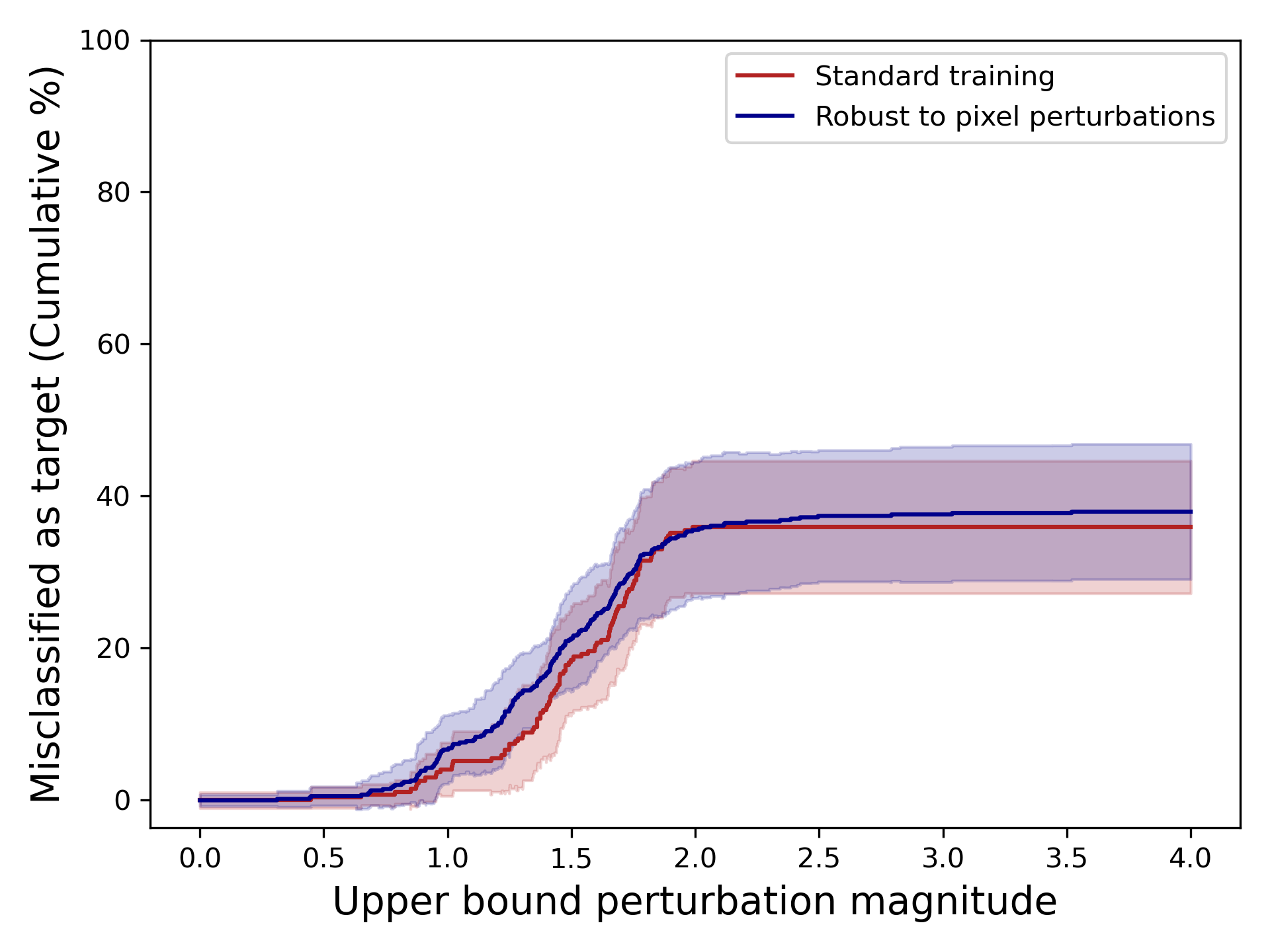}
		\caption{Generator activations perturbed at first 4 layers only.}
		\label{fig:mnist_graph_first}
	\end{subfigure}\hfill%
	\begin{subfigure}[t]{.49\textwidth}
		\centering
		\includegraphics[width=\linewidth]{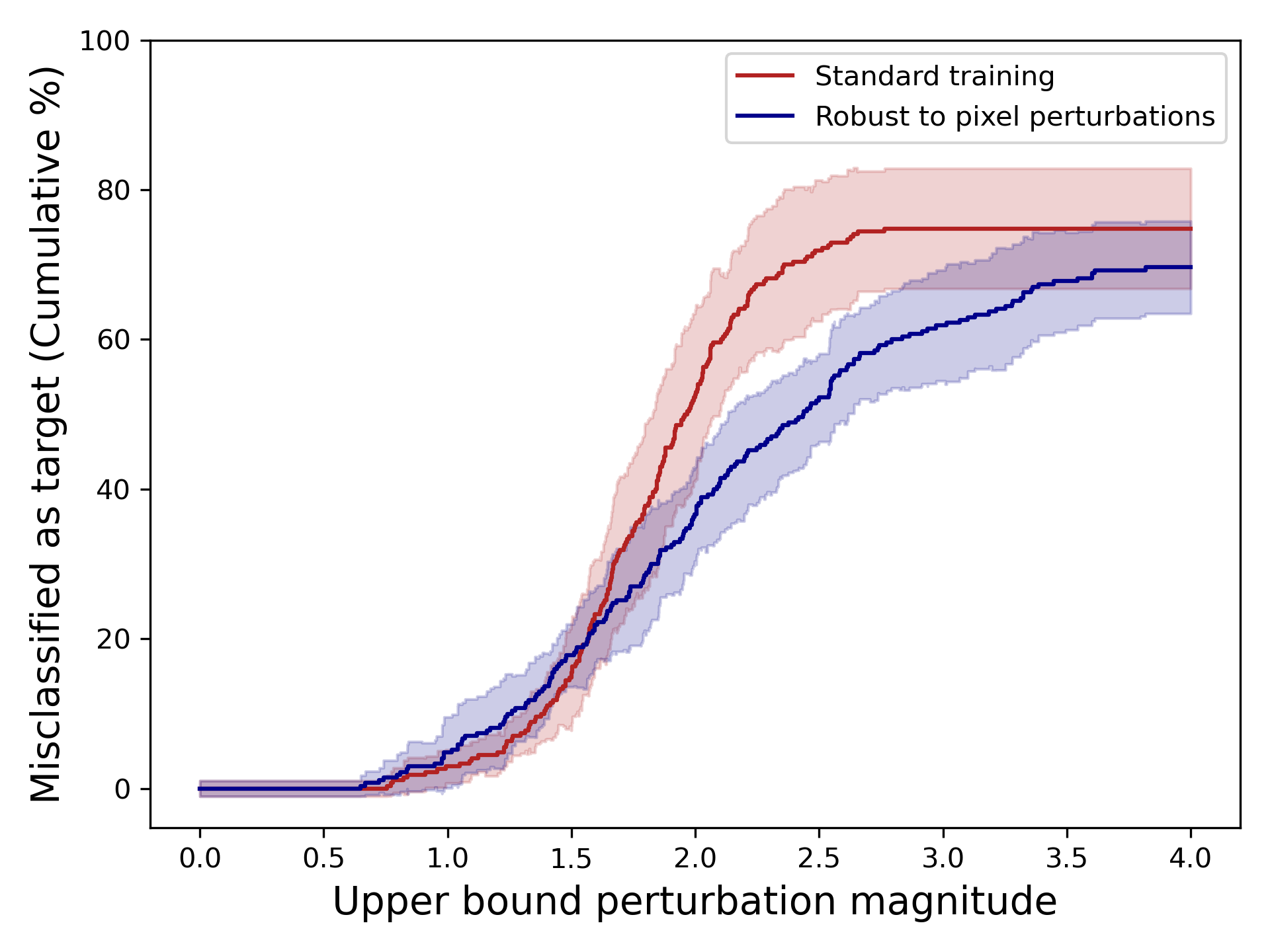}
		\caption{Generator activations perturbed at last 4 layers only.}
		\label{fig:mnist_graph_last}
	\end{subfigure}
	\caption{
	Graphs showing how the proportion of perturbations
	that induce the targeted misclassification
	increases with perturbation magnitude.
	These should be interpreted in the same way
	as Figure~\ref{fig:graphs}.
	The solid lines exclude the perturbed images for which
	a human judges that the perturbation does not change
	the true label of the image;
	the dotted lines, for reference, include these.}
	\label{fig:mnist_graphs}
\end{figure}

Figure~\ref{fig:mnist_graphs} shows the robustness of the
two classifiers to the two kinds of context-sensitive perturbation.
We can see from Figure~\ref{fig:mnist_graph_last} that
the classifier trained to be robust to pixel-space
perturbations is indeed more robust than the standard
classifier, with its considerably
shallower slope indicating that a bigger perturbation
magnitude is required to the finer-grained features
encoded in the last four layers of the generator.

Conversely, Figure~\ref{fig:mnist_graph_first}
gives an almost-identical shape for both classifiers,
indicating that adversarial training against pixel-space
perturbations does not confer any robustness to
the coarse-grained feature perturbations being
evaluated here.
This has an interesting relationship with our
main finding, that adversarial training against
pixel-space perturbations seriously harms robustness
to high-level context-sensitive features perturbations on ImageNet.
The difference can likely be best explained by the
great difference in datasets.
The dimensionality of ImageNet is
over $1000 \times$
greater, and the high-level feature space
of MNIST is trivially small in comparison.
The result of this is that in some loose sense,
there is a smaller `gap' between the
highest- and lowest-granularity features encoded
in the generator;
put another way, there is a much less rich space
of coarse-grained context-sensitive features that a classifier
must be robust to on MNIST.

Whether this is the correct intuition, the implications of
our finding remains clear: even on the very simplest datasets,
robustness to fine-grained features completely fails to generalise
to coarser-grained features.
If we are to obtain classifiers that we can trust to generalise
under modest distributional shifts, there is still far to go.

\newcommand{\mnistexample}[1]{\includegraphics[width=5mm]{images/mnist_results/#1_unpert.png}~\includegraphics[width=5mm]{images/mnist_results/#1.png} \hfill}

\newcommand{\mnistversion}[1]{
		\mnistexample{#1/0_0}
		\mnistexample{#1/0_1}
		\mnistexample{#1/0_2} 
		\mnistexample{#1/0_3} 
		\mnistexample{#1/0_4} 
		\mnistexample{#1/0_5} \\
		\mnistexample{#1/0_6} 
		\mnistexample{#1/0_7}
		\mnistexample{#1/0_8}
		\mnistexample{#1/0_9}
		\mnistexample{#1/0_10}
		\mnistexample{#1/0_11} \\
		\mnistexample{#1/0_12} 
		\mnistexample{#1/0_13} 
		\mnistexample{#1/0_14} 
		\mnistexample{#1/0_15}
		\mnistexample{#1/0_16} 
		\mnistexample{#1/0_17} \\
		\mnistexample{#1/0_18}
		\mnistexample{#1/0_19}
		\mnistexample{#1/0_20}
		\mnistexample{#1/0_21}
		\mnistexample{#1/0_22} 
		\mnistexample{#1/0_23} \\
}

\begin{figure}[p]
	\mnistversion{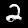}
	
	\caption{Random sample of context-sensitive perturbations targeting
		label 0. Only the first four layers of generator activations
		are perturbed, and the classifier is trained using a standard
		training procedure. In each pair, the perturbed image
		is to the right of the unperturbed image.}
\end{figure}

\begin{figure}[p]
	
	\mnistversion{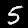}
	
	\caption{Random sample of context-sensitive perturbations targeting
		label 0. Only the first four layers of generator activations
		are perturbed, and the classifier is trained using adversarial
		training. In each pair, the perturbed image
		is to the right of the unperturbed image.}
\end{figure}

\begin{figure}[p]
	
	\mnistversion{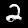}
	
	\caption{
	Random sample of context-sensitive perturbations targeting
	label 0. Only the last four layers of generator activations
	are perturbed, and the classifier is trained using a standard
	training procedure. In each pair, the perturbed image
	is to the right of the unperturbed image.
	}
\end{figure}

\begin{figure}[p]
	
	\mnistversion{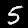}
	
	\caption{Random sample of context-sensitive perturbations targeting
		label 0. Only the last four layers of generator activations
		are perturbed, and the classifier is trained using adversarial
		training. In each pair, the perturbed image
		is to the right of the unperturbed image.}
\end{figure}

\end{document}